\documentclass[preprint,3p,times,onecolumn,authoryear]{elsarticle}

\usepackage{microtype}

\usepackage{graphicx}     
\usepackage{caption}      
\usepackage{subcaption}   
\usepackage{balance}
\usepackage{tabularx}
\usepackage{longtable}
\usepackage{url}
\usepackage{array}

\usepackage{soul}   

\usepackage{booktabs}
\newcolumntype{Y}{>{\centering\arraybackslash}X} 

\usepackage{multirow}
\usepackage{amssymb}      
\usepackage{amsmath,amsfonts}
\usepackage{textcomp}     

\usepackage[
colorlinks=true,
allcolors=blue
]{hyperref}               

\usepackage{float}

\usepackage{threeparttable} 
\usepackage{enumitem}
\newlist{tabitemize}{itemize}{1}
\setlist[tabitemize]{nosep,leftmargin=*}
\makeatletter
\newcommand{\compress}{\@minipagetrue} 
\makeatother

\usepackage{pgf}

\everymath=\expandafter{\the\everymath\displaystyle}
\makeatletter\@ifpackageloaded{underscore}{}{\usepackage[strings]{underscore}}\makeatother

\begin{document}

\begin{frontmatter}

  \title{Illicit object detection in X-ray imaging using deep learning techniques: A comparative evaluation}
  \author[1]{Jorgen Cani}
  \ead{cani@hua.gr}

  \author[1]{Christos Diou}
  \ead{cdiou@hua.gr}

  \author[2]{Spyridon Evangelatos}
  \ead{sevangelatos@netcompany.com}

  \author[3]{Vasileios Argyriou}
  \ead{Vasileios.Argyriou@kingston.ac.uk}

  \author[4,5]{Panagiotis Radoglou‐Grammatikis}
  \ead{pradoglou@k3y.bg}

  \author[4]{Panagiotis Sarigiannidis}
  \ead{psarigiannidis@uowm.gr}

  \author[1]{Iraklis Varlamis}
  \ead{varlamis@hua.gr}

  \author[1]{Georgios Th.~Papadopoulos}
  \ead{g.th.papadopoulos@hua.gr}

  \affiliation[1]{
    organization={Department of Informatics and Telematics, Harokopio University of Athens},
    addressline={Athens},
    country={Greece}
  }
  \affiliation[2]{
    organization={Research \& Innovation Development Department, Netcompany-Intrasoft S.A.},
    addressline={Luxembourg},
    country={Luxembourg}
  }
  \affiliation[3]{%
    organization={Department of Networks and Digital Media, Kingston University},
    addressline={London},
    country={United Kingdom}
  }
  \affiliation[4]{%
    organization={Department of Electrical and Computer Engineering, University of Western Macedonia},
    addressline={Kozani},
    country={Greece}
  }
  \affiliation[5]{%
    organization={K3Y Ltd},
    addressline={Sofia},
    country={Bulgaria}
  }

  \begin{abstract}
    Automated X-ray inspection is crucial for efficient and unobtrusive security screening in various public settings. However, challenges such as object occlusion/overlap, variations in the physical properties of the items of interest, diversity in the types of X-ray scanning devices used, and limited training data hinder accurate and reliable detection of illicit items. Despite the large body of research works in the field, the reported experimental evaluation is often incomplete, while the derived outcomes are frequently conflicting. In order to shed light on the research landscape of this field and to facilitate further research, a systematic, detailed, and thorough comparative evaluation study of recent Deep Learning (DL)-based methods for X-ray object detection is conducted in this work. For achieving this, a comprehensive evaluation framework is developed, composed of the following building blocks: a) Six of the most recent, large-scale and widely used public datasets for X-ray illicit item detection (namely, OPIXray, CLCXray, SIXray, EDS, HiXray, and PIDray), b) Ten different state-of-the-art object detection schemes, covering all main categories present in the literature, including generic Convolutional Neural Network (CNN), custom (X-ray-specific) CNN, generic transformer and generic hybrid CNN-transformer architectures, and c) Various detection (mAP\textsuperscript{50} and mAP\textsuperscript{50:95} mean Average Precision (mAP)) and time/computational-complexity (inference time (ms), parameter size (M), and computational load (GFLOPS)) performance metrics. A thorough analysis of the computed experimental results leads to the extraction of critical observations and detailed insights, emphasizing on the following key aspects: a) Overall behavior of the various object detection schemes, b) Object-level detection performance investigation, c) Dataset-specific observations, and d) Time efficiency and computational complexity analysis. In order to support reproducibility of the reported experimental results and to promote research in the field, the evaluation framework code and model weights are publicly available at \url{https://github.com/jgenc/xray-comparative-evaluation}.
  \end{abstract}

  \begin{keyword}
    X-ray imaging \sep object detection \sep convolutional neural networks \sep transformers \sep hybrid CNN-transformer architectures
  \end{keyword}

\end{frontmatter}

\section{Introduction}
\label{sec:Introduction}

Over the last decades, X-ray imaging has been established as the fundamental building block of inspection schemes in security-critical environments. In particular, non-destructive, unobtrusive, and harmless X-ray screening infrastructure is widely used in multiple security checkpoint locations (e.g., airports, customs, post offices, governmental buildings, stadiums, public event venues, etc.) to identify security threats (e.g., handguns, explosives, etc.) in trafficked packaging (e.g., parcels, baggage, containers, etc.). X-ray imaging relies on the use of high-energy electromagnetic radiation with wavelengths shorter than ultraviolet and longer than gamma rays. When such ion beams penetrate scanned objects, the X-ray signal is variably attenuated, depending on the mass density of the exposed objects. Consequently, the measured intensity of the output signal is inversely proportional to the density of the examined materials. This property is exploited by security services to efficiently analyze the packages' content and to identify possible threats, such as illicit or hazardous items \citep{partridge2022enhanced, kayalvizhi2022raw, mademlis2024invisible}.

X-ray imaging techniques can be roughly categorized based on two main criteria, namely the number of energy levels (of the X-ray beams) and the number of scanning views utilized \citep{velayudhan2022recent}. With regard to the number of employed energy levels, X-ray imaging can be divided into mono- and multi-energy level methods. Mono-energy X-rays use a single energy level of electromagnetic radiation to produce grayscale images, based on the mass density of the examined materials. In contrast, dual- and, more generally, multi-level energy X-ray scanners employ multiple energy levels and generate multi-channel X-ray images that enable a more detailed and high-quality representation of the material density. In order to facilitate the inspection process, the latter images are pseudo-colorized, using a look-up table that associates different colors to different material types \citep{abidi2005screener}. Concerning the number of scanning views, X-ray imaging can be split into 2D and 3D techniques. In the case of 2D imaging, X-rays penetrate the examined objects from a single direction; hence, producing a single 2D output image. Differently, in 3D imaging multiple axial slices are stacked into a single 3D representation/volume via post-processing, typically relying on Computed Tomography (CT) scanning techniques. It should be noted that 3D imaging, although providing richer information, is a significantly more time-consuming process compared to 2D analysis and requires substantially more expensive equipment. As a result, the vast majority of operational X-ray screening infrastructure relies on the use of 2D images (i.e., single-view capturing setups), typically involving multi-level energy scanners.

Despite the extensive usage of X-ray screening devices, the actual inspection process is still predominantly realized by human operators, relying on the experience, training, and knowledge capacity of the involved security staff. The latter fact though poses significant drawbacks and risks \citep{schwaninger2008impact, bolfing2008image, michel2007computer}, including, among others: a) The monotonous, stressful, and concentration-intensive nature of the task; b) Insufficient training procedures for operators; c) A considerable likelihood of human error, even with rigorous training programs; d) Susceptibility of the inspection process to factors such as fatigue, cognitive overload, emotional stress, and job dissatisfaction; and e) The inherently time-consuming nature of manual examination. Therefore, the development of automated, accurate, time-efficient, and robust solutions for packaging inspection becomes of paramount importance.

\begin{figure} [t]
  \centering
  \begin{tabular}{cc}
    \begin{tabular}{c}
      \begin{subfigure}[t]{0.2\textwidth}
        \includegraphics[width=\textwidth]{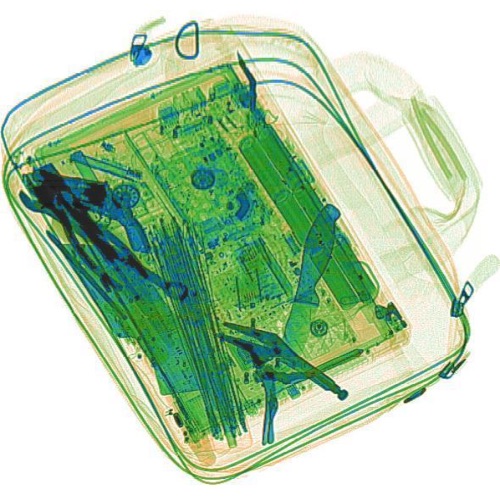}
        \caption{}
        \label{fig:sixray-sample}
      \end{subfigure}
    \end{tabular} &
    \begin{tabular}{c}
      \begin{subfigure}[t]{0.2\textwidth}
        \includegraphics[width=\textwidth]{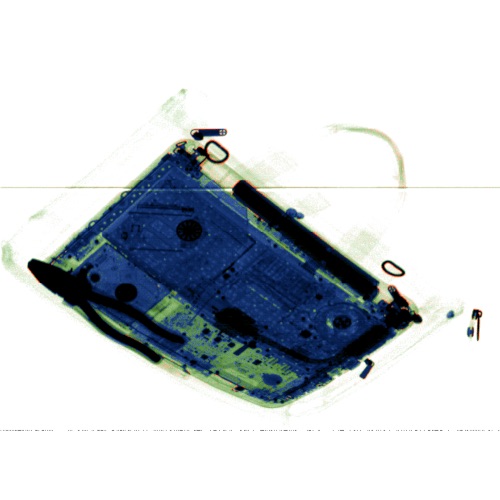}
        \caption{}
        \label{fig:pidray-sample}
      \end{subfigure}
    \end{tabular}
  \end{tabular}
  \caption{Indicative X-ray scan images from: a) The SIXray \citep{miaoSIXrayLargeScaleSecurity2019}, and b) The PIDray \citep{wangRealWorldProhibitedItem2021} datasets.}
  \label{fig:X-ray-samples}
\end{figure}

Towards automating the X-ray-based examination process, several conventional image processing/analysis and Machine Learning (ML)-based approaches have been investigated \citep{singh2024advancements}. Regardless of the particular methodology followed though, constructing robust automated threat detection systems faces several important challenges that include, among others \citep{velayudhan2022recent}: a) Lack of texture and poor contrast, inherently met in X-ray scans, b) Presence of extreme clutter, overlapping and (self-) occlusions, caused by the typically compact stacking of objects of varying material densities in an unstructured way (i.e. lack of orientation) in packages, c) Unavailability of sufficiently large (and annotated) datasets, mainly due to sensitivity and copyright issues in collecting security X-ray scans, d) Extreme class imbalance, caused by the rare observation of prohibited items in real-world security screening applications, e) Limited prior or expert knowledge, which relates to the natural uncertainty regarding the contents of a package, as well as their interpretation as a thread, f) Poor resolution and image quality, induced by the operational need for high scanning speed (over collecting high-quality imagery) and the presence of metal artefacts (that distort the captured X-ray images), g) Limited generalization ability, related to both the large variance in object appearance (high intra-class variance), as well as variations among technical specs across different (types/models of) scanners, and h) Existence of evolving threats, which is associated with the continuous need for adaptation to the introduction of new types of objects/threads or changes/evolution in the appearance of existing ones. Indicative examples of X-ray scan images showcasing some of the above-mentioned challenges are demonstrated in Fig. \ref{fig:X-ray-samples}.

Recent advances in Deep Learning (DL) \citep{alimisis2025advances, rodis2024multimodal}, combined with the introduction of larger 2D X-ray public datasets, have stimulated research and significantly contributed towards developing robust fully-automated inspection systems \citep{seyfi2024literature, rafiei2023computer}. With respect to the specific image analysis tasks considered, particular attention has been given to image classification, object detection, image segmentation, and anomaly detection \citep{wu2023object, gaikwad2025self}. Among the aforementioned categories, increased emphasis has been devoted on object detection techniques that are especially relevant and well-suited for threat identification applications. In particular, different types of Convolutional Neural Network (CNN) \citep{miaoSIXrayLargeScaleSecurity2019, weiOccludedProhibitedItems2020}, transformer \citep{liAODETRAntiOverlappingDETR2024, velayudhanTransformersImbalancedBaggage2022}, and hybrid CNN-transformer \citep{wuEslaXDETNewXray2024, ahmedDetectionTransformerFramework2023} methods have been proposed for detecting illicit/prohibited objects in 2D X-ray inspection images.

Despite the significant research efforts devoted and the important accomplishments reported in 2D X-ray object detection, further improvements need to be realized, in order to meet the operational needs of real-world inspection systems \citep{mery2020x}. Additionally, the landscape of this rapidly emerging/evolving research field exhibits several critical inconsistencies and limitations that need to be addressed, so as to boost further developments. In particular, critical concerns have already been highlighted in the relevant literature, including, among others \citep{mery2020x, rafiei2023computer, velayudhan2022recent}: a) The available object detection methods are typically evaluated using few (and often private) datasets, i.e., failing to realize robust and thorough performance analysis across a wide set of experimental settings, b) The reported experimental results are often not comparable across different studies (even for the same public datasets employed), due to variations/differences in the adopted experimental protocols/setups (e.g., data subsets, training/test set splitting, evaluation metrics, etc.), and c) The selected performance metrics are often not reported in details; common metrics, such as mean average precision, can have multiple implementations that need to be carefully described. The above observations suggest that a comprehensive and thorough comparative evaluation of the main methodological categories of recent X-ray-based object detection methods (i.e., CNN, transformer, and hybrid CNN-transformer approaches) across multiple/diverse datasets and using the exact same experimental protocols/metrics would greatly facilitate towards generating detailed/reliable observations/insights regarding the developments in the field and drawing promising future research directions.

In this paper, the problem of automatic (illicit) object detection in 2D (single-view) X-ray images using deep learning techniques is systematically investigated. In particular, the main contributions of this work are summarized as follows:
\begin{itemize}
  \item A comprehensive reporting of the publicly available datasets for X-ray-based packaging inspection is provided.
  \item A thorough analysis of the literature DL-based X-ray object detection methods, which are broadly categorized into generic CNN, custom (X-ray-specific) CNN, generic transformer and generic hybrid CNN-transformer approaches, is performed.
  \item Development of a comprehensive comparative evaluation framework, composed of the following main building blocks:
        \begin{itemize}
          \item Six of the most recent, large-scale and widely used public datasets for X-ray illicit item detection (namely, OPIXray, CLCXray, SIXray, EDS, HiXray, and PIDray),
          \item Ten different state-of-the-art object detection schemes, covering all above-mentioned main categories present in the literature,
          \item Various detection (mAP\textsuperscript{50} and mAP\textsuperscript{50:95} mean Average Precision (mAP)) and time/computational-complexity (inference time (ms), parameter size (M), and computational load (GFLOPS)) performance metrics.
        \end{itemize}
  \item Extraction of critical observations and detailed insights from the computed experimental results, emphasizing on the following key aspects: a) Overall behavior of the various object detection schemes, b) Object-level detection performance investigation, c) Dataset-specific observations, and d) Time efficiency and computational complexity analysis.
  \item In order to facilitate the reproducibility of the generated experimental results and to promote research in the field, the source code and the model weights of the developed evaluation framework are publicly available at \url{https://github.com/jgenc/xray-comparative-evaluation}
\end{itemize}

The remainder of the paper is organized as follows: Section \ref{sec:Datasets} details the publicly available datasets for X-ray-based packaging inspection. Section \ref{sec:RelatedWork} presents the recent literature on illicit object detection in X-ray scan images using DL techniques. Section \ref{sec:EvaluationFramework} outlines the defined experimental framework. Section \ref{sec:EvaluationResults} describes the computed comparative evaluation results, along with critical findings and insights that are observed. Section \ref{sec:Conclusions} concludes the study and discusses possible future research directions.

\section{Public datasets}
\label{sec:Datasets}

This section presents the publicly available datasets for X-ray-based packaging inspection. In contrast to the case of (general-purpose) RGB benchmarks, X-ray ones are in general relatively scarce, limited in size, and often tailored to specific computer vision tasks. Table \ref{tab:datasets} illustrates the datasets that are mostly used in the relevant literature along with their main characteristics, including: a) Name: Dataset name, b) Year: Publication or release year, c) Task: Specific tasks for which the dataset is designed, namely Multi-Label Classification (MLC), Object Localization (OL), Object Detection (OD), Few-Shot Object Detection (FSOD), Few-Shot Segmentation (FSS), Instance Segmentation (IS), Anomaly Detection (AD), and Image Classification (IC), d) Classes: Number of supported distinct object classes, e) Images: Total number of images, f) Annotation (abbreviated as `Annot.'): Type of available annotations, namely bounding box (bbox), segmentation mask (segm), and class label (cls), g) Color: Image color format, namely Grayscale (G) or RGB, h) Energy: X-ray beam energy levels, namely Single or Dual, and i) Description: Brief dataset description, including any notable features, characteristics, or additional information relevant to its use.

\begin{table}[!htbp]
  \footnotesize
  \renewcommand\tabularxcolumn[1]{m{#1}}    
  \centering
  \caption{Main public datasets for X-ray packaging inspection. An asterisk (*) indicates benchmarks used in the conducted comparative evaluation.}
  \begin{tabularx}{\textwidth}{ m{2cm} m{0.5cm} >{\centering\arraybackslash}m{0.7cm}  >{\centering\arraybackslash}m{0.9cm}  >{\centering\arraybackslash}m{1cm}  >{\centering\arraybackslash}m{0.8cm}  >{\centering\arraybackslash}m{0.7cm}  >{\centering\arraybackslash}m{0.8cm} X }
    \toprule
    \textbf{Name}                                         & \textbf{Year} & \textbf{Task} & \textbf{Classes} & \textbf{Images} & \textbf{Annot.} & \textbf{Color} & \textbf{Energy} & \textbf{Description}                                                                                                                       \\
    \midrule
    DET-COMPASS \citep{garcia2025superpowering}           & 2025          & OD            & 370              & 3,865           & bbox            & RGB            & Dual            & Pixel-aligned X-ray and RGB image pairs. Derived from the COMPASS-XP dataset.                                                              \\
    \midrule
    DVXray \citep{maDualViewXRayBaggage2024}              & 2024          & MLC, OL       & 15               & 32,000          & bbox            & RGB            & Dual            & Dual view image pairs per scan. Contains firearm, knife and metal categories.                                                              \\
    \midrule
    X-Adv \citep{liuXAdvPhysicalAdversarial2023}          & 2023          & OD            & 4                & 4,537           & bbox            & RGB            & Dual            & The test set contains adversarial images of concealed illicit objects, generated by taking into account the X-ray scanner characteristics. \\
    \midrule
    SIXray-D* \citep{nguyenMoreEfficientSecurity2022}     & 2022          & OD            & 6                & 11,401          & bbox            & RGB            & Dual            & Addition of bbox annotations to the original SIXray dataset. Update/correction of annotations from negative image set.                     \\
    \midrule
    CLCXray* \citep{zhaoDetectingOverlappedObjects2022}   & 2022          & OD            & 12               & 9,565           & bbox            & RGB            & Dual            & Inclusion of cutter and liquid objects, which are not present in other datasets.                                                           \\
    \midrule
    EDS* \citep{taoExploringEndogenousShift2022}          & 2022          & OD            & 10               & 14,219          & bbox            & RGB            & Dual            & Usage of three different scanners. Domain shift experimental protocol for evaluating models' transferability.                              \\
    \midrule
    FSOD \citep{taoFewshotXrayProhibited2022}             & 2022          & FSOD          & 20               & 12,333          & bbox            & RGB            & Dual            & 15 base classes used for training and 5 novel classes considered for evaluation.                                                           \\
    \midrule
    Xray-PI \citep{liuFewShotSegmentationProhibited2023}  & 2022          & FSS           & 7                & 2,409           & segm            & RGB            & Dual            & Firearm, knife, explosives, and everyday objects. 4 categories used for training and 3 novel classes considered for testing.               \\
    \midrule
    PIXray \citep{maAutomatedSegmentationProhibited2023}  & 2022          & IS, OD        & 15               & 5,046           & segm            & RGB            & Dual            & Segmentation-level annotations. Inclusion of non-metal objects. Increased overlap of depicted items. High-quality images.                  \\
    \midrule
    PIDray* \citep{wangRealWorldProhibitedItem2021}       & 2021          & OD            & 12               & 47,677          & bbox, segm      & RGB            & Dual            & Deliberately hidden images. Test set split into `easy', `hard', and `hidden'.                                                              \\
    \midrule
    HIXray* \citep{taoRealworldXraySecurity2021}          & 2021          & OD            & 8                & 45,364          & bbox            & RGB            & Dual            & High-quality X-ray images. Annotated by professional personnel. Everyday object classes (does not contain firearm or knife variants).      \\
    \midrule
    OPIXray* \citep{weiOccludedProhibitedItems2020}       & 2020          & OD            & 5                & 8,885           & bbox            & RGB            & Dual            & Evaluation of varying object occlusion levels. Mainly contains knife variants.                                                             \\
    \midrule
    COMPASS-XP \citep{caldwellLimitsTransferLearning2019} & 2020          & AD            & 369              & 1,901           & cls             & G, RGB         & Dual            & Few instances per object class. Low-level of (illicit) object occlusion.                                                                   \\
    \midrule
    SIXray \citep{miaoSIXrayLargeScaleSecurity2019}       & 2019          & MLC, OL       & 6                & 1,059,231       & cls             & RGB            & Dual            & Very high number of negative samples. Only 0.85\% of total images contain illicit objects.                                                 \\
    \midrule
    GDXray \citep{meryGDXrayDatabaseXray2015}             & 2015          & IC, OD        & 5                & 19,407          & bbox            & G              & Single          & Incorporation of castings, welds, baggage, and natural objects. Illicit items depicted both in baggage and standalone.                     \\
    \bottomrule
  \end{tabularx}
  \label{tab:datasets}
\end{table}

\section{DL-based object detection methods}
\label{sec:RelatedWork}

The aim of illicit object detection approaches in X-ray packaging inspection is to determine both the class and the location of each identified thread within an image, typically in the form of an axis-aligned rectangular bounding box. Taking into account the type of the employed Neural Network (NN) architecture, literature approaches can broadly be classified into generic CNN, custom (X-ray-specific) CNN, generic transformer, and generic hybrid CNN-transformer methods, as described in Section \ref{sec:Introduction}.

Table \ref{tab:od-methods} demonstrates key and best-performing methods of the literature (organized according to their adopted NN architecture type), along with their main characteristics, including: a) Method: Method name, b) Year: Publication year, c) Task: Specific task(s) for which the method is designed, namely Object Detection (OD), Open Vocabulary Object Detection (OV-OD), Segmentation (S), Few-Shot Object Detection (FS-OD), Classification (C), and Zero-Shot Classification (ZS-C), d) Detector type: The primary object detector framework utilized by the method, namely R-CNN, YOLO, multiple, or custom, e) Base detection network: Base NN architecture utilized for performing object detection, f) Backbone: Backbone NN architecture utilized for extracting visual features from the input image, g) Learning strategy: NN learning strategy adopted during training, namely Supervised (S), Weakly Supervised (WS), Adversarial (A), Meta-Transfer learning (MT), Distillation-based supervised learning (D), and Few-Shot learning (FS), and h) Code: Public release status of the method's implementation (a checkmark ($\checkmark$) indicates if source code is available, a dagger on top of the checkmark ($\checkmark^{\dagger}$) if both code and pretrained weights are provided, or a dash (--) if no code is available). In the remaining of the section, the different categories of DL-based object detection methods for X-ray packaging inspection are discussed in details.

\subsection{Generic CNN methods}
\label{ssec:GenericCNNMethods}
Following the successful application of CNNs to object detection in conventional RGB images \citep{sultana2020review}, these networks have also been widely used for identifying threats in X-ray imagery. In the followings, the relevant literature of generic CNN approaches is systematically analyzed, taking into account the type of the detection scheme adopted, namely Region-based Convolutional Neural Network (R-CNN)-based, You Only Look Once (YOLO)-based, and incorporation of multiple detectors.

\subsubsection{R-CNN-based detectors}
R-CNN \citep{girshick2014rich} and its subsequent variants/extensions (e.g., Faster R-CNN \citep{renFasterRCNNRealTime2015a}) are among the first CNN architectures used for object detection and have also been extensively used for X-ray image analysis. In particular, the Selective Dense Attention Network (SDANet) \citep{wangRealWorldProhibitedItem2021} employs selective channel-wise and spatial attention modules to enhance object detection and segmentation, utilizing a dense attention mechanism and a dependency refinement module to account for multi-scale features. The Material-aware Cross-channel Interaction Attention (MCIA) module \citep{wangMaterialawareCrosschannelInteraction2023} uses material data in X-ray images to tackle inter-class occlusions, by integrating into Residual Network (ResNet) stages. MCIA includes a Material Perception (MP) and a Cross-channel Interaction (CI) component, which emphasize prohibited items and suppress non-prohibited ones, improving detection in Faster R-CNN and Cascade R-CNN models. Additionally, the Perturbation Suppression Network (PSN) \citep{taoExploringEndogenousShift2022} addresses endogenous shift for cross-domain detection using Local Prototype Alignment (LPA) and Global Adversarial Assimilation (GAA) to mitigate category-dependent disruptions. In parallel, the Weak-feature Enhancement Network (WEN) \citep{taoFewshotXrayProhibited2022} enhances few-shot object detection, using prototype perception and feature reconciliation mechanisms to improve feature distinctiveness through iterative prototype updates. The MAM Faster R-CNN model \citep{zhangMAMFasterRCNN2023} introduces a Malformed Attention Module (MAM) to expand the convolutional receptive field of the feature map and to extract local features of objects with shape distortion, using a Large Kernel Attention (LKA) block and a Path Aggregation Network (PAN) for enhancing feature focus. Moreover, an end-to-end Weakly Supervised Correction (WSC) approach is presented in \citep{wangLearningAmbiguousLabels2024}, in order to denoise and to rectify ambiguous labels, featuring X-ray Energy Awareness Blending (X-Blending), a Weakly Supervised Head (WSH) and an Adaptive Label Corrector (ALC) to generate credible labels and to adjust sample contributions.

\begin{table*}[!htbp]
  \centering
  \caption{DL-based object detection methods for X-ray packaging inspection. An asterisk (*) indicates approaches used in the conducted comparative evaluation. A checkmark (\checkmark) indicates available source code, a dagger on top of the checkmark (\checkmark$^\dagger$) indicates that both code and pretrained weights are provided, and a dash (--) indicates that no code is available.}
  \scriptsize
  \renewcommand\tabularxcolumn[1]{m{#1}}
  \begin{tabularx}{\textwidth}{
    m{3cm}
    >{\centering\arraybackslash}m{0.3cm}
    >{\centering\arraybackslash}m{0.7cm}
    >{\centering\arraybackslash}m{0.8cm}
    >{\centering\arraybackslash}m{3.3cm}
    >{\centering\arraybackslash}m{3.5cm}
    >{\centering\arraybackslash}m{0.9cm}
    >{\centering\arraybackslash}m{0.5cm}
    }
    \toprule
    \textbf{Method}                                                  & \textbf{Year} & \textbf{Task} & \textbf{Detector type} & \textbf{Base detection network}           & \textbf{Backbone}                                                                           & \textbf{Learning strategy} & \textbf{Code}          \\
    \midrule
    \multicolumn{8}{l}{\textbf{Generic CNN}}                                                                                                                                                                                                                                                                                  \\
    \midrule
    TinyRay     \citep{zhangLightweightProhibitedItems2025}          & 2025          & OD            & YOLO                   & Custom YOLOv7-tiny                        & FasterNet                                                                                   & S                          & \checkmark$^{\dagger}$ \\
    \midrule
    XFKD   \citep{renFeatureKnowledgeDistillationbased2025}          & 2025          & OD            & Multiple               & RetinaNet, YOLOv4                         & CSPDarkNet-\{53, 23\}, ResNet-\{50,101\}, MobileNetV3, DenseNet, GhostNet, and ShuffleNetV2 & S                          & \checkmark             \\
    \midrule
    YOLO-SRW \citep{chen2025yolo}                                    & 2025          & OD            & YOLO                   & Custom YOLOv8                             & Customized CSPDarkNet53                                                                     & S                          & --                     \\
    \midrule
    \cite{chen2025x}                                                 & 2025          & OD            & YOLO                   & Custom YOLOv8s                            & Default Backbone                                                                            & S                          & --                     \\

    \midrule
    \cite{wangLightweightDetectionMethod2024}                        & 2024          & OD            & YOLO                   & Custom YOLOv8-n                           & Default backbone                                                                            & S                          & --                     \\
    \midrule
    X-YOLO \citep{cheng2024x}                                        & 2024          & OD            & YOLO                   & Custom YOLOv5s                            & CSPDarkNet53                                                                                & S                          & --                     \\
    \midrule
    YOLOV8-n-GEMA   \citep{wangImprovedYOLOv8Dangerous2024}          & 2024          & OD            & YOLO                   & Custom YOLOv8-n                           & Default backbone                                                                            & S                          & --                     \\
    \midrule
    YOLOV8s-DCN-EMA-IPIO \citep{gaoContrabandDetectionScheme2024}    & 2024          & OD            & YOLO                   & Custom YOLOv8-n                           & Customized CSPDarkNet53                                                                     & S                          & --                     \\
    \midrule
    WSC      \citep{wangLearningAmbiguousLabels2024}                 & 2024          & OD            & R-CNN                  & Faster R-CNN                              & ResNet-50-FPN                                                                               & S, WS                      & --                     \\
    \midrule
    SC-Lite     \citep{hanSCLiteEfficientLightweight2024}            & 2024          & OD            & YOLO                   & Custom YOLOv8                             & Customized CSPDarkNet53                                                                     & S                          & --                     \\
    \midrule
    \cite{batsis2023illicit}                                         & 2023          & OD            & YOLO                   & YOLOv5                                    & CSPDarkNet53                                                                                & S                          & --                     \\
    \midrule
    POD      \citep{maOccludedProhibitedObject2023}                  & 2023          & OD            & Multiple               & Faster R-CNN, YOLOv5L                     & ResNet-50, ResNeXt-50, CSPDarkNet-53                                                        & S                          & \checkmark             \\
    \midrule
    EM-YOLO    \citep{jingEMYOLOXrayProhibitedItemDetection2023}     & 2023          & OD            & YOLO                   & YOLOv7                                    & ResNet, DenseNet                                                                            & S                          & --                     \\
    \midrule
    SC-YOLOv8   \citep{hanSCYOLOv8SecurityCheck2023}                 & 2023          & OD            & YOLO                   & Custom YOLOv8                             & Customized CSPDarkNet53                                                                     & S                          & --                     \\
    \midrule
    MAM Faster R-CNN \citep{zhangMAMFasterRCNN2023}                  & 2023          & OD            & R-CNN                  & Faster R-CNN                              & ResNet-50                                                                                   & S                          & --                     \\
    \midrule
    MCIA-Net    \citep{wangMaterialawareCrosschannelInteraction2023} & 2023          & OD            & R-CNN                  & Faster R-CNN, Cascade R-CNN               & ResNet-101                                                                                  & S                          & --                     \\
    \midrule
    WEN      \citep{taoFewshotXrayProhibited2022}                    & 2022          & FS-OD         & R-CNN                  & Faster R-CNN                              & ResNet-101                                                                                  & S                          & \checkmark             \\
    \midrule
    PSN       \citep{taoExploringEndogenousShift2022}                & 2022          & OD            & R-CNN                  & Faster R-CNN                              & VGG-16                                                                                      & S, A                       & \checkmark             \\

    \midrule
    SDANet     \citep{wangRealWorldProhibitedItem2021}               & 2021          & OD            & R-CNN                  & Cascade Mask-RCNN                         & ResNet-101                                                                                  & S                          & \checkmark$^{\dagger}$ \\

    \midrule

    \multicolumn{8}{l}{\textbf{Custom (X-ray-specific) CNN}}
    \\

    \midrule
    FDTNet    \citep{zhuFDTNetEnhancingFrequencyaware2024}           & 2024          & OD            & Custom                 & Custom                                    & ResNeXt101                                                                                  & S                          & --                     \\

    \midrule
    CPID \citep{wang2024delving}                                     & 2024          & OD, S         & Multiple               & Faster R-CNN, Mask R-CNN, Cascade R-CNN   & ResNet-101                                                                                  & S                          & --                     \\

    \midrule
    DDoAS    \citep{maAutomatedSegmentationProhibited2023}           & 2022          & S, OD         & Custom                 & Customized DeepSnake                      & ResNet-50, VGG-16, Inception-v3, Densenet-121                                               & S                          & \checkmark$^{\dagger}$ \\

    \midrule
    LA        \citep{zhaoDetectingOverlappedObjects2022}             & 2022          & OD            & Custom                 & ATSS                                      & ResNet-50                                                                                   & S                          & \checkmark$^{\dagger}$ \\

    \midrule
    TDC         \citep{nguyenMoreEfficientSecurity2022}              & 2022          & OD            & Custom                 & RFB-Net                                   & RFB-Net                                                                                     & S                          & --                     \\

    \midrule
    CFPA-Net     \citep{weiCFPANetCrosslayerFeature2021}             & 2021          & C, OD         & Custom                 & RetinaNet                                 & ResNet                                                                                      & S                          & --                     \\

    \midrule
    LIM*       \citep{taoRealworldXraySecurity2021}                  & 2021          & OD            & Multiple               & SSD, FCOS, YOLOv5                         & VGG16, ResNet-50, CSPNet                                                                    & S                          & \checkmark$^{\dagger}$ \\

    \midrule
    CST       \citep{hassanMetaTransferLearningDriven2020}           & 2020          & OD, ZS-C      & Custom                 & Custom                                    & ResNet-\{50, 101\}, VGG-16                                                                  & MT                         & --                     \\

    \midrule
    DOAM*      \citep{weiOccludedProhibitedItems2020}                & 2020          & OD            & Multiple               & SSD, YOLOv3, FCOS                         & VGG16, DarkNet-53, ResNet-50                                                                & S                          & \checkmark$^{\dagger}$ \\

    \midrule

    \multicolumn{8}{l}{\textbf{Generic transformer}}                                                                                                                                                                                                                                                                          \\
    \midrule

    MHT-X \citep{alansari2024multi}                                  & 2024          & OD            & Custom                 & Custom transformer-based                  & Custom ViT                                                                                  & S                          & --                     \\
    \midrule
    AO-DETR \citep{liAODETRAntiOverlappingDETR2024}                  & 2024          & OD            & DINO                   & DINO                                      & ResNet-50, Swin-L                                                                           & S                          & \checkmark             \\
    \midrule

    \multicolumn{8}{l}{\textbf{Generic hybrid CNN-transformer}}                                                                                                                                                                                                                                                               \\
    \midrule

    \cite{cani2025BDS}                                               & 2025          & OD            & Custom                 & Custom YOLOv8 and RT-DETR                 & HGNetV2, Next-ViT-S                                                                         & S                          & \checkmark$^{\dagger}$ \\
    \midrule

    DGDN \citep{yang2025novel}                                       & 2025          & OD            & Custom                 & Custom hybrid CNN- and Vision Mamba-based & ResNet-50, Modified CSP-DarkNet53                                                           & S                          & --                     \\
    \midrule

    OVXD  \citep{linDetectionNovelProhibited2025}                    & 2025          & OV-OD         & R-CNN                  & Faster R-CNN                              & ResNet-50                                                                                   & D                          & --                     \\
    \midrule
    Xray-YOLO-Mamba \citep{zhao2025lightweight}                      & 2025          & OD            & YOLO                   & YOLOv11-n                                 & Custom Vision Mamba-based                                                                   & S                          & --                     \\
    \midrule

    MSFA-DETR \citep{sima2024multi}                                  & 2024          & OD            & Custom                 & Custom DETR-based                         & ResNet-50                                                                                   & S                          & --                     \\
    \midrule

    Trans2ray  \citep{mengTransformerbasedDualviewXray2024}          & 2024          & OD            & Custom                 & Custom Transformer-based                  & ResNet-50                                                                                   & S                          & \checkmark             \\
    \midrule
    AdaptXray   \citep{huangAdaptxrayVisionTransformer2024}          & 2024          & OD            & Multiple               & ViT-Det with Cascade R-CNN head           & ViT-B                                                                                       & S                          & --                     \\
    \midrule
    EslaXDET     \citep{wuEslaXDETNewXray2024}                       & 2024          & OD            & R-CNN                  & Cascade Mask R-CNN                        & ViT-B                                                                                       & S                          & --                     \\
    \midrule
    BGM    \citep{liuBGMBackgroundMixup2024}                         & 2024          & OD            & Multiple               & Both CNNs and DINO                        & -                                                                                           & S                          & --                     \\
    \midrule
    RVViT   \citep{liuNewFewShotLearningBased2023}                   & 2023          & FS-OD         & Multiple               & Custom hybrid CNN- and transformer-based  & ResNet-50                                                                                   & FS                         & --                     \\
    \midrule
    \cite{ahmedDetectionTransformerFramework2023}                    & 2023          & OD            & Custom                 & Customized DETR                           & ResNet-50                                                                                   & S                          & --                     \\

    \bottomrule
  \end{tabularx}
  \label{tab:od-methods}
\end{table*}

\subsubsection{YOLO-based detectors}
The more recent `You Only Look Once' (YOLO) model and its multiple subsequent versions \citep{vijayakumar2024yolo} have also been extensively used in X-ray object detection. In particular, EM-YOLO \citep{jingEMYOLOXrayProhibitedItemDetection2023} employs two pre-processing modules before utilizing a modified YOLOv7, namely an Edge Feature Extraction (EFE) (inspired by DOAM \citep{weiOccludedProhibitedItems2020}) and a Material Feature Extraction (MFE) one. The SC-YOLOv8 \citep{hanSCYOLOv8SecurityCheck2023} model introduces a CSPnet Deformable Convolution Network Module (C2F\_DCN) and a Spatial Pyramid Multi-Head Attention Module (SPMA), in order to enhance feature representations across different scales. Additionally, YOLOv8n-GEMA \citep{wangImprovedYOLOv8Dangerous2024} employs a Generalized Efficient Layer Aggregation Network (GELAN) and an Efficient Multi-Scale Attention (EMA) scheme, in order to address overlap and occlusion occurrences. \cite{wangLightweightDetectionMethod2024} propose a YOLOv8-based method that combines an Adaptive Spatial Feature Fusion (ASFF) and a Coordinate Attention (CoordAtt) module, aiming to enhance feature learning and to handle occlusions. In parallel, SC-Lite \citep{hanSCLiteEfficientLightweight2024} is designed for real-time detection in resource-limited environments, incorporating a CSPNet Faster Convolution Network Module (C2F\_FM) and an Adaptation-BiFPN one for optimal feature fusion. Moreover, TinyRay \citep{zhangLightweightProhibitedItems2025} enhances YOLOv7-tiny with a lightweight FasterNet backbone and a New-ELAN module, in order to optimize resource usage. Likewise, YOLO-SRW \citep{chen2025yolo} modifies YOLOv8 to dynamically adjust spatial receptive fields, using the RFLSKA module for multi-scale feature extraction; its Wise-SIoU loss incorporates angular information and balances sample quality, reducing errors and improving generalization. \cite{batsis2023illicit} enhance YOLOv5 using Hierarchical Clustering (HC) for anchor box generation, aligning with ground-truth object size and shape distributions across classes; a Weighted Cluster Non-Maximum Suppression (WC-NMS) scheme is applied to manage complexity and an Efficient-IoU (E-IoU) metric for modeling detailed geometrical information. Similarly, \cite{chen2025x} integrate a Multi-scale Cross-axis Attention (MCA) module into YOLOv8 to capture global dependencies, using Partial Convolution (PConv) to create a more efficient bottleneck architecture and a Focaler-IoU loss function to enhance regression accuracy on difficult samples. Furthermore, the X-YOLO model \citep{cheng2024x} incorporates a Soft Convolutional Block Attention Module (Soft-CBAM), which incorporates a SoftPool operator to better retain sub-pixel information and an improved dynamic head module to unify feature attention across different scales and tasks.

\subsubsection{Multiple detectors}
In an attempt to achieve increased recognition performance, while maintaining general applicability, a series of methods have evaluated multiple CNN detectors, including different versions of YOLO, Faster R-CNN, and others. In particular, the De-Occlusion Attention Module (DOAM) \citep{weiOccludedProhibitedItems2020}, which is used in combination with various detection methods (namely, SSD, YOLOv3, and FCOS), incorporates edge and material information, in order to create an attention map that preserves target shapes under occlusion. Additionally, the Lateral Inhibition Module (LIM) \citep{taoRealworldXraySecurity2021}, which is evaluated using common detection approaches (namely, SSD, FCOS, and YOLOv5), aims to reduce noise and to enhance object boundaries, through the employment of bidirectional propagation and boundary activation mechanisms. Moreover, the Prohibited Object Detection (POD) method \citep{maOccludedProhibitedObject2023} employs a Gabor convolutional layer for edge extraction, a Spatial Attention (SA) mechanism for structure enhancement, a Global Context Feature Extraction (GCFE) module for estimating multi-scale global contextual information and a Dual Scale Feature Aggregation (DSFA) module for performing feature fusion; the aforementioned modules are embedded into the Faster R-CNN and YOLOv5L object detection frameworks. Furthermore, the XFKD \citep{renFeatureKnowledgeDistillationbased2025} approach combines a Local Distillation (LD) and a Global Distillation (GD) mechanism to improve lightweight models' performance, while being evaluated using RetinaNet and YOLOv4.

\subsection{Custom CNN detectors}
\label{ssec:CustomCNNMethods}
Apart from adopting common generic CNN-based detection schemes (e.g., R-CNN, YOLO, SSD, etc.), additional custom-designed object detectors or approaches originally developed for other/similar X-ray image analysis tasks (e.g., image segmentation), which can however been used for performing object detection, have been proposed. In particular, the Class-balanced Hierarchical Refinement (CHR) module \citep{miaoSIXrayLargeScaleSecurity2019} refines features hierarchically and eliminates irrelevant information, aiming at improving classification in imbalanced datasets. Additionally, the Cascaded Structure Tensor (CST) method \citep{hassanMetaTransferLearningDriven2020} processes low- and high-energy tensors to extract contour-based proposals, while being integrated with pre-trained networks (like ResNet and DenseNet). The Security X-ray Multi-label Classification Network (SXMNet) \citep{huMultilabelXRayImagery2021} incorporates a ResNet50-FPN backbone and an attention head, realizing feature refinement for generating the final predictions using a meta fusion scheme. In parallel, the Cross-layer feature Fusion and Parallel Attention network (CFPA-Net) \citep{weiCFPANetCrosslayerFeature2021} enhances RetinaNet by integrating three modules, namely a Cross-layer feature Extraction Fusion module (CEF-Module), a Paralleled Attention Module (PA-Module) and the FreeAnchor one, in order to emphasize task-related object features. The Task-Driven Cropping (TDC) scheme \citep{nguyenMoreEfficientSecurity2022} removes unnecessary background from X-ray scans, in an attempt to enhance the detection performance. Moreover, the Label-Aware mechanism (LA) \citep{zhaoDetectingOverlappedObjects2022} aims to address the object overlapping problem, by establishing associations between feature channels and different labels, and adjusting the features according to the assigned labels. Furthermore, for addressing data scarcity and improving feature representation for cluttered items, the Cluttered Prohibited Item Detection (CPID) \citep{wang2024delving} method combines an online random cut-and-paste data augmentation strategy with a High-Order Dilated Convolution (HDC) module, which is designed to enrich feature discriminability and to enlarge the receptive field.

Concerning approaches primarily developed for tasks other than object detection, the Dense De-overlap Attention Snake (DDoaS) method \citep{maAutomatedSegmentationProhibited2023} is designed for real-time prohibited item segmentation, aiming at efficiently handling overlapping items. Additionally, a patch-based self-supervised learning method, combined with a Prototype Reverse Validation strategy (PRV) \citep{liuFewShotSegmentationProhibited2023}, is adopted for few-shot prohibited items segmentation, leveraging unlabeled data to learn abstract representations. Additionally, the dual-stream frequency-aware detection network (FDTNet) \citep{zhuFDTNetEnhancingFrequencyaware2024} enhances prohibited item representation using frequency domain information, while being capable of being integrated into various backbones or detectors. Moreover, the Adaptive Hierarchical Cross Refinement (AHCR) method \citep{maDualViewXRayBaggage2024} comprises a multi-view architecture analyzing dual-view X-ray images, fusing features from both views in order to enhance discrimination ability.

\subsection{Generic transformer methods}
\label{ssec:TransformerMethods}
Vision Transformer (ViT) \citep{han2022survey}, i.e., a particular type of NN architecture introduced more recently than the CNN one, has also showcased outstanding performance in various RGB image analysis tasks. However, its requirement for increased amounts of training data (compared to CNNs), combined with the unavailability of large public X-ray benchmarks, has hindered its wide adoption for threat identification. Nevertheless, the recent introduction of sizable public X-ray datasets has encouraged the development of transformer-based packaging inspection schemes. In particular, \cite{velayudhanTransformersImbalancedBaggage2022} explore the usage of vision transformers for imbalanced baggage threat recognition, leveraging their ability to model global features, in order to capture concealed illicit items within cluttered and tightly packed baggage scans. Additionally, the Anti-Overlapping DEtection TRansformer (AO-DETR) \citep{liAODETRAntiOverlappingDETR2024} integrates a Category-Specific Assignment (CSA) strategy into the DINO framework, aligning category-oriented queries with reference boxes for reducing overlap confusion. Moreover, MHT-X \citep{alansari2024multi} leverages ViT to address occlusion and clutter with multi-scale contour mapping; it incorporates a spatial reduction block within its transformer encoder for hierarchical information.

\subsection{Generic hybrid CNN-transformer methods}
\label{ssec:HybridMethods}

In an attempt to further increase object detection performance, hybrid network architectures have also been introduced, which combine transformer (for capturing long-range dependencies) and CNN (for extracting local information) building blocks \citep{guo2022cmt, khan2023survey}, often with additional sophisticated architectural and learning components. Reasonably, such composite NN architectures have been incorporated in X-ray imaging inspection systems. In particular, the Trans2Ray \citep{mengTransformerbasedDualviewXray2024} method relies on the use of a dual-view vision transformer that incorporates two channels. The main channel is responsible for the detection of prohibited objects, while the secondary one provides valuable features to enhance the main channel; feature extraction in both cases is performed using a ResNet-50 backbone. \cite{linDetectionNovelProhibited2025} introduce an Open-Vocabulary X-ray prohibited item Detection (OVXD) model, which extends CLIP to learn visual representations in the X-ray domain, aiming to detect novel prohibited item categories beyond the base ones. Additionally, \cite{garcia2025superpowering} proposed RAXO, a training-free framework that adapts off-the-shelf RGB open-vocabulary detectors for X-ray vision by constructing robust visual descriptors from web-retrieved and in-domain images, achieving superior performance without the need for retraining. \cite{ahmedDetectionTransformerFramework2023} propose a DEtection TRansformer (DETR) framework, which relies on receiving extracted features from a CNN backbone, using object proposals derived from coherent contour maps. The RVViT \citep{liuNewFewShotLearningBased2023} method enhances the stability of the few-shot learning paradigm, by adopting a transformer encoder for generating high-level semantic features that contain global information, while also devising an edge detection module for boosting the edge information of prohibited items. Moreover, EslaXDET \citep{wuEslaXDETNewXray2024} combines a backbone, trained using a hybrid Self-Supervised Learning (SSL) strategy \citep{konstantakos2025self}, and a detection head, which creates multi-level feature maps, by down-sampling multiple times the output feature of the last stage of the plain ViT. AdaptXray \citep{huangAdaptxrayVisionTransformer2024} utilizes a pre-trained vision transformer with a parameter efficient transfer learning scheme. In parallel, the Background Mixup (BGM) \citep{liuBGMBackgroundMixup2024} method introduces a patch-level data augmentation approach, combining baggage contour and material variation information. In order to better handle object size disparities, the Multi-Scale Feature Attention DETR (MSFA-DETR) \citep{sima2024multi} method embeds a pyramid feature structure built with atrous convolutions into the self-attention module, while a foreground sequence extraction module improves the initialization of object queries to speed up convergence. Xray-YOLO-Mamba \citep{zhao2025lightweight} is a lightweight model merging YOLO and VMamba \citep{liu2024vmamba} for efficient X-ray image analysis. It utilizes specialized blocks CResVSS, SDConv, and Dysample to enhance feature representation and resolution. Furthermore, the Dangerous Goods Detection Network (DGDN) \citep{yang2025novel} architecture pairs a purely CNN-based channel adaptive module with a hybrid spatial adaptive module that leverages the Mamba module principles to refine spatial features; this hybrid approach allows the model to effectively handle overlapping goods and to suppress irrelevant background noise. \cite{cani2025BDS} introduce various hybrid CNN-transformer architectures; more specifically, a CNN (HGNetV2) and a hybrid (Next-ViT-S) backbone are combined with different CNN/transformer detection heads (YOLOv8 and RT-DETR).

\section{Comparative evaluation framework}
\label{sec:EvaluationFramework}

\begin{table*}[!htbp]
  \renewcommand\tabularxcolumn[1]{m{#1}}
  \centering
  \caption{Building blocks of the comparative evaluation framework for X-ray prohibited object detection performance assessment.}
  \scriptsize
  \begin{tabularx}{\textwidth}{
      m{1.2cm} X X X X X
    }
    \toprule
     & \textbf{Public datasets}                                                                                                                                                                                                                                                                                        & \textbf{Object detection heads} & \textbf{Backbone networks} & \textbf{Performance metrics} & \textbf{Utilized implementations} \\
    \midrule
    \textbf{Options considered}
     & OPIXray \citep{weiOccludedProhibitedItems2020}, CLCXray \citep{zhaoDetectingOverlappedObjects2022}, SIXray \citep{miaoSIXrayLargeScaleSecurity2019}, EDS \citep{taoExploringEndogenousShift2022}, HiXray \citep{taoRealworldXraySecurity2021}, PIDray \citep{wangRealWorldProhibitedItem2021}
     & YOLOv8 \citep{jocherUltralyticsYOLOv82023}, CHR \citep{miaoSIXrayLargeScaleSecurity2019}, DOAM \citep{weiOccludedProhibitedItems2020}, LIM \citep{taoRealworldXraySecurity2021}, DINO \citep{zhangDINODETRImproved2022}, Co-DETR \citep{zongDETRsCollaborativeHybrid2023}, RT-DETR \citep{lvDETRsBeatYOLOs2023}
     & CSPDarkNet53 \citep{wang2020cspnet}, HGNetV2 \citep{HGNetv2}, Swin-B \citep{liuSwinTransformerHierarchical2021}, Next-ViT-S \citep{liNextViTNextGeneration2022}
     & Object detection (mAP\textsuperscript{50}, mAP\textsuperscript{50:95}), time/computational-complexity (inference time (ms), parameter size (M), computational load (GFLOPS))
     & Authors' publicly available code, public toolboxes (Ultralytics, MMDetection)
    \\
    \midrule
    \textbf{Rationale}
     & Thorough evaluation under varying experimental settings, with respect to object types, item sizes, degree of occlusion, level of clutter/complexity, dataset size, capturing setup, etc.
     & Examination of the behavior of various state-of-art object detection heads, including generic CNN, custom CNN and generic transformer ones.
     & Investigation of the behavior of various state-of-art backbone networks for feature extraction, including CNN, transformer, and hybrid ones.
     & Simultaneous evaluation of both detection and time performance for practical/operational deployment assessment.
     & Assurance of experiments' reproducibility and evaluation transparency.
    \\
    \bottomrule
  \end{tabularx}
  \label{tab:comparative-framework}
\end{table*}

This section outlines the defined comparative evaluation framework, which is used in the current study for thoroughly and comprehensively assessing the behavior/performance of the various types of DL-based object detection methods for X-ray packaging inspection (as detailed in Section \ref{sec:RelatedWork}). In particular, the main constituting components and selections of the developed framework, which are briefly summarized in Table \ref{tab:comparative-framework} and further discussed below, comprise: a) Public datasets/benchmarks for X-ray experimental assessment, b) Object detection heads, including generic CNN (YOLOv8), custom (X-ray-specific) CNN (CHR, DOAM, LIM) and generic transformer (DINO, Co-DETR, RT-DETR) ones, c) Backbone networks, including CNN (CSPDarkNet53, HGNetV2), transformer (Swin-B) and hybrid CNN-transformer (Next-ViT-S) ones, d) Performance metrics, including both detection (mAP\textsuperscript{50} and mAP\textsuperscript{50:95} mean Average Precision (mAP)) and time/computational-complexity (inference time (ms), parameter size (M), and computational load (GFLOPS)) ones, and e) Method implementation details. It needs to be highlighted that the source code and network weights of all models included in the evaluation framework are available at \url{https://github.com/jgenc/xray-comparative-evaluation}.

\subsection{Datasets}
\label{ssec:EF-datasets}

\begin{figure*}[!htbp]
  \centering
  \begin{tabular}{c}

    \begin{tabular}{c c c c}
      \begin{subfigure}[t]{0.2\textwidth}
        \includegraphics[width=\textwidth]{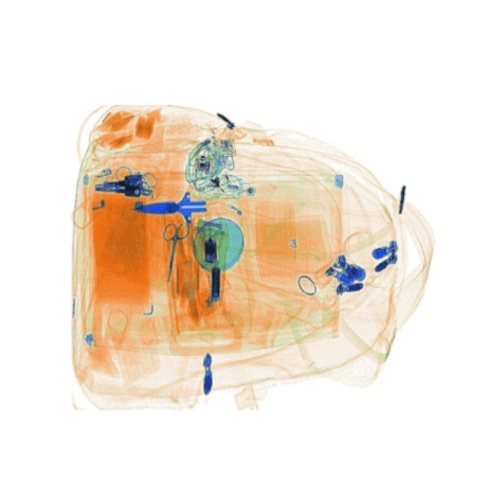}
        \caption{}
        \label{fig:opixray-sample}
      \end{subfigure} &
      \begin{subfigure}[t]{0.2\textwidth}
        \includegraphics[width=\textwidth]{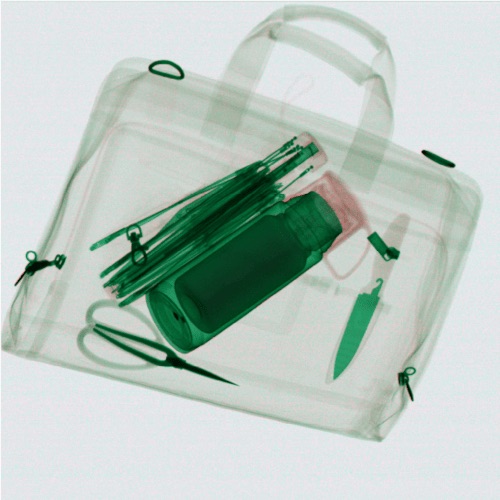}
        \caption{}
        \label{fig:clcxray-sample}
      \end{subfigure} &
      \begin{subfigure}[t]{0.2\textwidth}
        \includegraphics[width=\textwidth]{Images/sixray_fig.jpeg}
        \caption{}
        \label{fig:sixray-sample}
      \end{subfigure}  &
      \begin{subfigure}[t]{0.2\textwidth}
        \includegraphics[width=\textwidth]{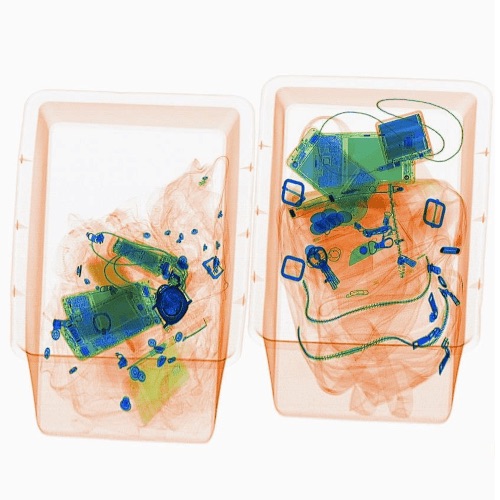}
        \caption{}
        \label{fig:hixray-sample}
      \end{subfigure}  \\
    \end{tabular}

    \vspace{0.1cm}

    \\

    \begin{tabular}{c c}
      \begin{subfigure}[t]{0.2\textwidth}
        \includegraphics[width=\textwidth]{Images/pidray_fig.jpeg}
        \caption{}
        \label{fig:pidray-sample}
      \end{subfigure} ~~~~ & ~~~~
      \begin{subfigure}[t]{0.55\textwidth}
        \includegraphics[width=\textwidth]{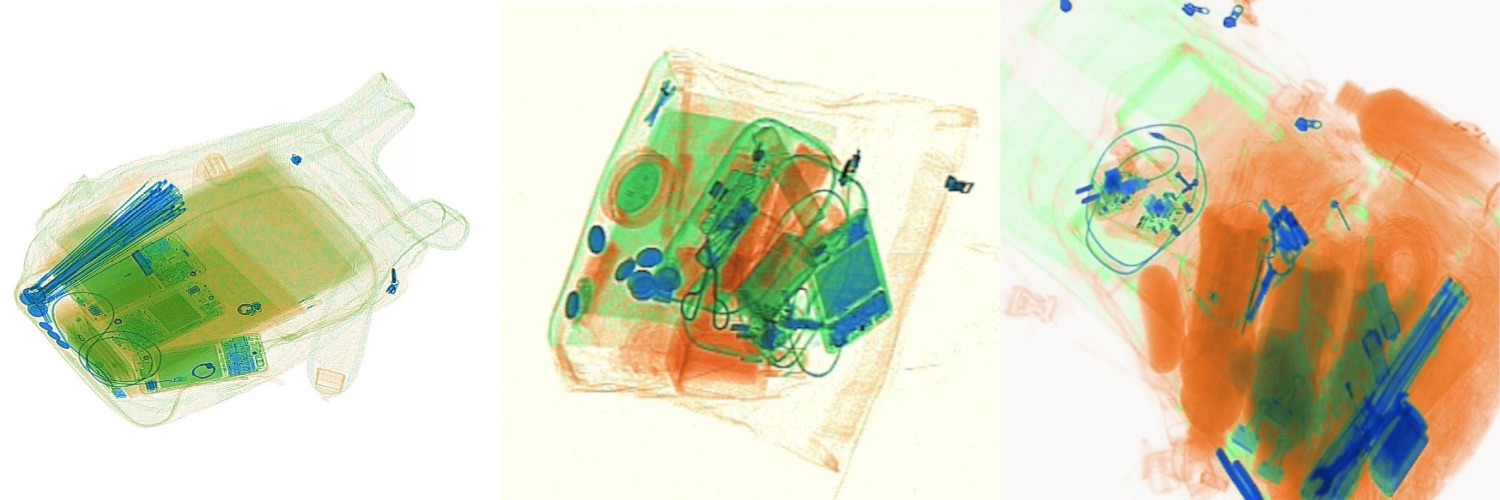}
        \caption{}
        \label{fig:eds-sample}
      \end{subfigure}
    \end{tabular}
  \end{tabular}

  \caption{Exemplary images from the a) OPIXray \citep{weiOccludedProhibitedItems2020}, b)  CLCXray \citep{zhaoDetectingOverlappedObjects2022}, c) SIXray \citep{miaoSIXrayLargeScaleSecurity2019}, d) HiXray \citep{taoRealworldXraySecurity2021}, e) PIDray \citep{wangRealWorldProhibitedItem2021}, and f) EDS \citep{taoExploringEndogenousShift2022} datasets.}
  \label{fig:used-dataset-samples}
\end{figure*}

In order to ensure comprehensive and robust evaluation, across different experimental settings, multiple X-ray object detection benchmarks have been considered in this study. In particular, six of the most recent, large-scale and widely used public datasets have been employed, as indicated in Table \ref{tab:datasets} and further detailed below. Specifically, the utilized datasets are OPIXray \citep{weiOccludedProhibitedItems2020}, CLCXray \citep{zhaoDetectingOverlappedObjects2022}, SIXray \citep{miaoSIXrayLargeScaleSecurity2019}, EDS \citep{taoExploringEndogenousShift2022}, HiXray \citep{taoRealworldXraySecurity2021}, and PIDray \citep{wangRealWorldProhibitedItem2021}, while exemplary images of each of them are illustrated in Fig. \ref{fig:used-dataset-samples}.

The \textbf{OPIXray} \citep{weiOccludedProhibitedItems2020} dataset pays particular attention on investigating the issue of object occlusion in X-ray scans. In particular, real-world inspection scenarios typically involve the examination of objects that are positioned one on top of the others in the investigated package, resulting into varying levels of occlusion. In this context, for the creation of OPIXray, security inspectors were asked to simulated such real-world investigation cases, resulting in a set of X-ray scans that exhibit varying degrees of object occlusion. The simulation scenario took place at an international airport, focusing on the scanning of personal luggage and security bins. Five classes of prohibited items are considered, namely \textit{Folding knife (FO)}, \textit{Straight knife (ST)}, \textit{Scissor (SC)}, \textit{Utility knife (UT)}, and \textit{Multi-tool knife (MU)}. The dataset comprises a total of $8,885$ images, each containing at least one prohibited item that is annotated using a bounding box. OPIXray is divided into training and test sets, with the former comprising $80\%$ of the total images ($7,109$) and the latter containing the remaining $20\%$ ($1,776$). The test set is further split into three subsets, each associated with a different level of occlusion: OL1: Featuring items with no or slight occlusion, OL2: Showcasing partial occlusion occurrences, and OL3: Comprising images that are either severely or fully occluded.

The \textbf{CLCXray} \citep{zhaoDetectingOverlappedObjects2022} dataset also emphasizes on investigating the item occlusion challenge, concerning object overlaps with same-class instances as well as with their surrounding background. The dataset was created to address common limitations of existing X-ray security image benchmarks, which often lack sufficient overlap between multiple objects and neglect liquid containers. CLCXray contains a total of $9,565$ images, out of which $4,543$ are real samples (obtained from subway scan inspection systems) and $5,022$ are simulated instances (generated through scanning of artificially designed baggage setups). The dataset includes a total of twelve classes, belonging to two broad categories, namely cutters (\textit{Blade (BL)}, \textit{Dagger (DA)}, \textit{Knife (KN)}, \textit{Scissors (SC)}, and \textit{Swiss army knife (SW)}), and liquid containers (\textit{Can (CA)}, \textit{Carton drink (CD)}, \textit{Glass bottle (GB)}, \textit{Plastic bottle (PL)}, \textit{Vacuum cup (VA)}, \textit{Spray can (SP)}, and \textit{Tin (TI)}). CLCXray is split into three sets: a training ($80\%$ of images), a validation ($10\%$ of images), and a test ($10\%$ of images) one. It is noteworthy that the test set has been formed using an $1:9$ real-to-simulated sample ratio, whereas the respective ratio for the training and validation sets is equal to $8:1$. On average, each X-ray image contains more than two potentially dangerous items and nearly $60\%$ of X-ray images contain at least two or more foreground objects.

The \textbf{SIXray} \citep{miaoSIXrayLargeScaleSecurity2019} dataset comprises a substantial collection of X-rays images, including a total number of $1,059,231$ samples. From the aforementioned set, only $8,929$ images contain an object considered as prohibited, i.e., positive samples. The images were collected from multiple subway stations. The initial study examined the distribution of these images, as it reflects real-world situations where positive samples are significantly less numerous than negative ones. Although initially six classes were considered (including the under-represented \textit{Hammer} category), the following five classes are supported: \textit{Gun (GU)}, \textit{Knife (KN)}, \textit{Wrench (WR)}, \textit{Pliers (PL)}, and \textit{Scissors (SC)}. A notable extension of the SIXray dataset comprises the so-called \textbf{SIXray-D} \citep{nguyenMoreEfficientSecurity2022} one. Specifically, SIXray-D is created on top of the original SIXray, by considering a more efficient cropping scheme that enables the identification of additional positive samples within the $\sim1$ million negative images of SIXray. Following manual inspection and verification, a total of $2,578$ new positive images have been incorporated in SIXray-D. As a result, SIXray-D contains extra and more accurate annotations compared to SIXray, while each positive sample in SIXray-D is now determined using bounding box information. The authors have randomly divided the data into a training set, comprising $90\%$ of the images, and a test set, comprising $10\%$ of the images. This division is also utilized in this work. In order to avoid confusions, in the remaining of the manuscript the SIXray-D dataset is considered, although for simplicity the term SIXray is used.

The \textbf{EDS} \citep{taoExploringEndogenousShift2022} dataset focuses on the challenge of domain shift that is inherent in X-ray imaging, due to factors like varying parameters across different scanning devices. In particular, three different X-ray scanners are employed, resulting into variations in the captured color, depth and texture information channels, mainly introduced by the different device specs and wear levels. The packages used during the scanning process were artificially prepared by the authors. EDS supports ten classes of common daily-life objects, namely \textit{Plastic bottle (DB)}, \textit{Pressure (PR)}, \textit{Lighter (LI)}, \textit{Knife (KN)}, \textit{Device (SE)}, \textit{Power bank (PB)}, \textit{Umbrella (UM)}, \textit{Glass bottle (GB)}, \textit{Scissor (SC)}, and \textit{Laptop (LA)}. The dataset comprises $14,219$ images containing $31,654$ object instances from three domains (X-ray machines), resulting into $\sim2.22$ instances per image on average. The defined experimental protocol dictates the training of a detection model in a single domain and its subsequent evaluation in a different one, resulting in a total of six performed experimental sessions.

The \textbf{HiXray} \citep{taoRealworldXraySecurity2021} dataset contains real-world X-ray scans collected from an international airport, where the image annotations were provided by the airport security personnel. The dataset comprises $45,364$ images that include a total of $102,928$ prohibited items, i.e. $\sim2.27$ instances per image. The dataset supports eight classes, namely \textit{Portable charger 1 (lithium-ion prismatic cell) (PO1)}, \textit{Portable charger 2 (lithium-ion cylindrical cell) (PO2)}, \textit{Water (WA)}, \textit{Laptop (LA)}, \textit{Mobile phone (MP)}, \textit{Tablet (TA)}, \textit{Cosmetic (CO)}, and \textit{Nonmetallic Lighter (NL)}. HiXray is split into a training ($80\%$ of images) and a test ($20\%$ of images) set.

The \textbf{PIDray} \citep{wangRealWorldProhibitedItem2021} dataset focuses on deliberately hidden items, mimicking real-world scenarios where prohibited objects are intentionally concealed. The latter fact adds an extra level of complexity to the object detection task, since it is required to identify hidden items (and not `simply' detecting objects obscured by other items and/or environmental factors). All scan samples are collected under real-world settings, namely at airport, subway, and railway station security checkpoints. PIDray includes twelve classes of prohibited items, namely \textit{Gun (GU)}, \textit{Knife (KN)}, \textit{Wrench (WR)}, \textit{Pliers (PL)}, \textit{Scissors (SC)}, \textit{Hammer (HA)}, \textit{Handcuffs (HC)}, \textit{Baton (BA)}, \textit{Sprayer (SP)}, \textit{Power-bank (PB)}, \textit{Lighter (LI)}, and \textit{Bullet (BU)}. The dataset is split into a training ($29,457$ samples, $\sim60\%$ of images) and a test ($18,220$ samples, $\sim40\%$ of images) set. Moreover, the test set is further divided into three sub-sets, namely an easy (the images contain only one prohibited object), a hard (the images contain more than one illicit items), and a hidden (the images contain deliberately hidden objects) one, with $9,482$, $3,733$, and $5,055$ images, respectively.

\subsection{Object detection heads}
\label{ssec:EF-heads}

So far, a wide set of DL-based object detection methods has been introduced for X-ray package inspection (Section \ref{sec:RelatedWork}), which can be broadly categorized into generic CNN, custom CNN, generic transformer and generic hybrid approaches, based on the type of the employed NN architecture. In order to comparatively evaluate and assess the merits of each architectural building block, a broad set of the most recent, best-performing and widely used detection heads (that realize the actual prediction step, i.e., estimation of the bounding-box and the class scores for each identified object) and backbone networks (that extract feature representations from the input image, often in multiple resolutions) are investigated in the current study.

In the followings, the detection heads considered in the defined evaluation framework are outlined, including generic CNN (YOLOv8 \citep{jocherUltralyticsYOLOv82023}), custom CNN (CHR \citep{miaoSIXrayLargeScaleSecurity2019}, DOAM \citep{weiOccludedProhibitedItems2020}, LIM \citep{taoRealworldXraySecurity2021}), and generic transformer (DINO \citep{zhangDINODETRImproved2022}, Co-DETR \citep{zongDETRsCollaborativeHybrid2023}, RT-DETR \citep{lvDETRsBeatYOLOs2023}) network topologies. Critical characteristic of all selected detection heads, apart from their demonstrated recognition performance, is the availability of publicly available implementation/code. These detection heads are combined with various common backbone networks (Section \ref{ssec:EF-backbones}), forming multiple object detectors (Section \ref{ssec:EF-detectors}) to be comparatively evaluated in this study (Section \ref{sec:EvaluationResults}).

\subsubsection{Generic CNN detection heads}
\label{sssec:EF-CNN-heads}

The \textbf{YOLOv8} \citep{jocherUltralyticsYOLOv82023} method belongs to the so-called `You Only Look Once' (YOLO) series/family of methods \citep{redmonYouOnlyLook2015} and it is particularly designed for real-time application settings. YOLOv8 builds upon YOLOv5 \citep{yolov5}, though incorporating several key enhancements. Notably, YOLOv8 integrates an anchor-free detection head, which facilitates towards higher accuracy and more efficient detection performance, compared to anchor-based approaches. Additionally, it pays particular focus on maintaining an optimal balance between accuracy and speed. The YOLOv8 detection head employs multiple modules that predict bounding boxes, objectness scores, and class probabilities for each grid cell in the feature map; these predictions are subsequently aggregated to obtain the final detection. Among the various YOLOv8 model variants available, the YOLOv8l detection head is used in the current study, mainly due to computational resources availability aspects. Originally, YOLOv8 uses a custom CSPDarknet53 backbone, which employs cross-stage partial connections to improve information flow between layers and to boost accuracy. It needs to be highlighted that more recent versions of the YOLO approach have also been evaluated (namely, YOLOv12 \citep{tian2025yolov12}); however, it was experimentally shown to lead to negligible performance variations compared to YOLOv8. To this end, YOLOv8 is used in the current study, which also corresponds to the YOLO model most widely used in the relevant X-ray object detection literature.

\subsubsection{Custom CNN detection heads}
\label{sssec:EF-custom-CNN-heads}

The `Class-balanced Hierarchical Refinement' (\textbf{CHR}) \citep{miaoSIXrayLargeScaleSecurity2019} approach assumes that each X-ray image is sampled from a mixture distribution and that deep networks require an iterative process to infer image contents accurately. In order to accelerate this process, reversed connections are inserted to different network backbones, delivering high-level visual cues to assist mid-level features. Additionally, a class-balanced loss function is used to maximally alleviate noise introduced by easy negative samples. CHR can be combined with any CNN-based detection head. Originally, CHR is evaluated/combined with five different backbone networks, namely ResNet34, ResNet50, ResNet101, Inception-v3, and DenseNet121.

The `De-Occlusion Attention Module' (\textbf{DOAM}) \citep{weiOccludedProhibitedItems2020} approach pays particular attention on handing the item occlusion problem in X-ray images, relying on the fundamental principle that shape appearance of objects can be preserved at a satisfactory level. In particular, DOAM simultaneously leverages the varying appearance information of a prohibited item to generate an attention map, which facilitates the refinement of feature maps generated from generic object detectors. The latter is realized by laying particular emphasis on edge and material information of the prohibited items, as inspired by the X-ray imaging principle. DOAM can be combined with any generic CNN object detector. Originally, DOAM is evaluated/combined with the following CNN-based detectors: SSD, YOLOv3, and FCOS.

The `Lateral Inhibition Module' (\textbf{LIM})  \citep{taoRealworldXraySecurity2021} approach is inspired by the fact that humans recognize prohibited items in X-ray images, by ignoring irrelevant information and focusing on identifiable characteristics; especially, when objects are overlapping with each other. In particular, LIM suppresses noisy information flowing maximumly, making use of a bidirectional propagation mechanism, and activates the most identifiable boundary locations. LIM can be combined with any generic CNN-based object detector. Originally, LIM is evaluated/combined with the following CNN-based backbone networks: VGG16, ResNet50, and CSPNet.

\subsubsection{Generic transformer detection heads}
\label{sssec:EF-VIT-heads}

The `DETR with Improved deNoising anchOr boxes' (\textbf{DINO}) \citep{zhangDINODETRImproved2022} object detection method builts upon the DETR model \citep{lvDETRsBeatYOLOs2023}, incorporating, though, several key advantageous characteristics: a) It adopts a contrastive-based methodology for denoising training, b) It incorporates a mixed query selection approach for anchor initialization, and c) It integrates a look forward twice scheme for box prediction. In this way, DINO is proven superior to the original DETR model, both in terms of performance and efficiency. DINO incorporates a multi-head prediction mechanism that is considered in the current work. Originally, DINO is evaluated/combined with a transformer (Swin-L) backbone, as well as with a CNN (ResNet-50) one.

\textbf{Co-DETR} \citep{zongDETRsCollaborativeHybrid2023} incorporates a collaborative hybrid assignment training scheme, in order to learn more efficient and effective DETR-based detectors from versatile label assignments. In particular, this training scheme relies on the usage of multiple parallel auxiliary heads, supervised by one-to-many label assignments. Additionally, extra customized positive queries are conducted, by extracting the positive coordinates from these auxiliary heads to improve the training efficiency of positive samples in the decoder. During inference, the auxiliary heads are discarded. In this way, Co-DETR eventually relies on the DINO \citep{zhangDINODETRImproved2022} head topology. Originally, Co-DETR is evaluated/combined with three different backbone networks, namely ResNet-50, Swin-L, and ViT-L.

The `Real-Time DEtection TRansformer' (\textbf{RT-DETR}) \citep{lvDETRsBeatYOLOs2023} method enhances the DETR model, so as to produce a real-time end-to-end transformer-based object detector. In particular, RT-DETR incorporates an efficient hybrid encoder to expeditiously process multi-scale features, by decoupling intra-scale inter-action and cross-scale fusion to improve speed. Additionally, an uncertainty-minimal query selection approach is adopted to provide
high-quality initial queries to the decoder, in order to improve accuracy. RT-DETR incorporates a transformer decoder with auxiliary prediction heads that is employed in the current work. Originally, RT-DETR is evaluated/combined with two different backbone networks, namely ResNet50, and ResNet101 \citep{he2016deep}.

\subsection{Backbone networks}
\label{ssec:EF-backbones}

In the followings, the backbone modules considered in the defined evaluation framework are outlined, including CNN (CSPDarknet53 \citep{wang2020cspnet}, HGNetv2 \citep{HGNetv2}), transformer (Swin-B \citep{liuSwinTransformerHierarchical2021}), and hybrid (Next-ViT-S \citep{liNextViTNextGeneration2022}) network topologies. The aforementioned backbone networks were selected taking into account: a) Their demonstrated ability in generating discriminant feature representations, b) The availability of efficient and publicly available implementation/code, and c) The available computational resources utilized in this study.

The `Cross Stage Partial Network' (CSPNet) \citep{wang2020cspnet} model, also most commonly termed \textbf{CSPDarknet53}, is a CNN module that aims to mitigate the problem of requiring heavy inference computations from the network architecture perspective. The latter problem is mainly attributed to the duplicate gradient information considered within the network optimization procedures. To this end, CSPDarknet53 respects the gradient variability, by integrating feature maps from the beginning and the end of a network stage. This architectural design is experimentally shown to reduce computations and to lead to equivalent or even superior recognition performance.

The `High Performance GPU Network V2' (\textbf{HGNetv2}) \citep{HGNetv2} is a CNN high-performing backbone network that is more suitable for GPU accelerators. HGNetv2 relies on the use of a learnable down-sampling layer and a relatively simple semi-supervised knowledge distillation scheme. Additionally, HGNetv2 incorporates a learnable affine block module, which can facilitate towards improving recognition performance, while introducing few extra parameters. Moreover, its stage distribution is constructed to cover models of different orders of magnitude, so as to meet the needs of different analysis tasks.

The `Swin transformer' \citep{liuSwinTransformerHierarchical2021} comprises a hierarchical transformer architecture, whose representation is computed using shifted windows. This shifted windowing scheme results into greater efficiency by limiting self-attention computation to non-overlapping local windows, while also allowing for cross-window connection modeling. Its hierarchical architecture enables the network's flexibility to model features at various scales and has linear computational complexity with respect to image size. Out of the available architectural variants (namely Swin-T, Swin-S, Swin-B, and Swin-L, ordered according to increasing network size), the \textbf{Swin-B} backbone is considered in this work.

The `Next-ViT' \citep{liNextViTNextGeneration2022} model comprises a hybrid architecture targeting the efficient deployment in realistic industrial scenarios, aiming at optimizing the latency/accuracy trade-off. In particular, a Next Convolution Block (NCB) and a Next Transformer Block (NTB) are introduced to capture local and global information, respectively, exhibiting also deployment-friendly mechanisms. Then, a Next Hybrid Strategy (NHS) is introduced to stack NCB and NTB in an efficient hybrid paradigm, in order to enhance recognition performance. Out of the available architectural variants (namely Next-ViT-S, Next-ViT-B, and Next-ViT-L, ordered according to increasing network size), the \textbf{Next-ViT-S} backbone is considered in this work.

\subsection{Object detectors}
\label{ssec:EF-detectors}

DL-based X-ray object detection methods can broadly be classified into generic CNN, custom CNN, generic transformer, and hybrid CNN-transformer ones, taking into account the type of the employed NN architecture (Section \ref{sec:RelatedWork}). Since the fundamental goal of this study is to provide a comprehensive, thorough and detailed comparative evaluation of the various categories of approaches present in the literature, multiple combinations of detection heads (Section \ref{ssec:EF-heads}) and backbone networks (Section \ref{ssec:EF-backbones}) are considered, where each selected combination is denoted D(head, backbone). In particular, the following classes of object detectors are taken into account, accompanied with the corresponding motivation/justification behind each choice:

\begin{itemize}
  \item \underline{Generic CNN detectors}: The current literature for X-ray object detection is dominated by the use of adapted CNN methods, originally designed for conventional RGB analysis (Section \ref{ssec:GenericCNNMethods}). However, the following facts hold: a) In most cases the most recent CNN detection heads and backbone networks are not considered, and b) When publicly available implementations exist, these do not correspond to the most modern and powerful CNN architectures (Table \ref{tab:od-methods}). In this context, one of the most contemporary generic CNN object detection methods (YOLOv8) with its default backbone (CSPDarknet53) has been incorporated in the comparative evaluation study, forming detector \textbf{D(YOLOv8, CSPDarknet53)}. Additionally, a variant of the aforementioned model is also considered, by replacing the default backbone (CSPDarknet53) with the more recent HGNetV2; hence, forming detector \textbf{D(YOLOv8, HGNetV2)}.

  \item \underline{Custom CNN detectors}: As described in Section \ref{ssec:CustomCNNMethods}, a significant part of CNN methods for X-ray packaging inspection rely on the use of a customized CNN architecture. In this respect, the approaches of CHR \citep{miaoSIXrayLargeScaleSecurity2019}, DOAM \citep{weiOccludedProhibitedItems2020}, and LIM  \citep{taoRealworldXraySecurity2021} have shown outstanding performance, can be combined with any CNN architecture and provide publicly available implementations. To this end, these are combined with YOLOv8 and its default backbone (CSPDarknet53) in the current experimental study, forming detectors \textbf{D(YOLOv8+CHR, CSPDarknet53)}, \textbf{D(YOLOv8+DOAM, CSPDarknet53)}, and \textbf{D(YOLOv8+LIM, CSPDarknet53)}.

  \item \underline{Generic transformer detectors}: So far, end-to-end transformer architectures have received decreased attention in X-ray packaging inspection schemes (Section \ref{ssec:TransformerMethods}). In order to quantitatively investigate the behavior of transformer methods, combinations of some of the most recent and best performing modules are included in the comparative evaluation study, namely detectors \textbf{D(DINO, Swin-B)} and \textbf{D(Co-DETR, Swin-B)}.

  \item \underline{Generic hybrid CNN-transformer detectors}: Although hybrid CNN-transformer architectures have recently been introduced in the field of X-ray object detection, no sufficient/extensive experimental evaluation or publicly available implementations exist. In this context, various detection schemes (incorporating different/recent detection heads and backbone networks) are incorporated in this study, namely detectors \textbf{D(RT-DETR, HGNetv2)}, \textbf{D(YOLOv8, Next-ViT-S)}, and \textbf{D(RT-DETR, Next-ViT-S)}.
\end{itemize}

\subsection{Performance metrics}
\label{ssec:EF-metrics}

This section outlines the performance metrics used in the defined comparative evaluation framework for X-ray object detection, which include both object detection (mAP\textsuperscript{50} and mAP\textsuperscript{50:95} mean Average Precision (mAP)) and time/computational-complexity (inference time (ms), parameter size (M), and computational load (GFLOPS)) ones.

\subsubsection{Object detection metrics}

Average Precision (AP) and Mean Average Precision (mAP) constitute two of the most commonly used metrics in object detection applications \citep{padilla2020survey}, which estimate a comprehensive evaluation of the examined model's performance across different confidence levels and object classes. In general, higher AP and mAP scores indicate better performance. However, the definition/estimation of AP and mAP can vary slightly across different challenges and benchmarks. The particular definitions considered in this study are described in the followings.

Object detectors typically output a bounding box for each identified object, along with a confidence value for the respective predicted class. Examining each detected object separately, the Intersection over Union (IoU) metric assesses the spatial overlap between the predicted bounding box (generated by a detector model) and the corresponding ground truth one (that defines the actual location of the object). A high IoU score suggests that the model has not only correctly identified the object's class, but it has also accurately identified its location within the examined image. The calculation of the IoU score involves the determination of the area of intersection between the two examined bounding boxes (predicted and ground truth), divided by the area of their union. Given two bounding boxes $A$ and $B$, the IoU metric is calculated as follows:
\begin{equation}
  \text{IoU} = \frac{|A \cap B|}{|A \cup B|} \in [0,1]
\end{equation}
Typically, a minimum threshold value $T$ is considered for the IoU score (degree of overlap), so as to assess the respective detection as valid/correct.

Having identified detections using IoU, the precision and recall metrics provide complementary performance insights, analyzing the accuracy and completeness of the detections, respectively. In particular, precision measures the accuracy of the positive predictions and it is calculated as the ratio of True Positives (TP) to the total number of predictions (i.e., true positives plus False Positives (FP)), as follows:
\begin{equation}
  \text{Precision}=\frac{\text{TP}}{\text{TP}+\text{FP}}
\end{equation}
A high precision value indicates that the examined model avoids false positive predictions. On the other hand, recall measures the ability of the model to find all relevant objects present in an image and it is calculated as the ratio of TP to the total number of actual objects (i.e., TP plus False Negatives (FN)), as follows:
\begin{equation}
  \text{Recall}=\frac{\text{TP}}{\text{TP}+\text{FN}}
\end{equation}
A high recall value indicates that the examined model is efficient in identifying most of the actual objects.

Average Precision (AP) is a metric that provides a more comprehensive evaluation of a model's performance, compared to precision or recall alone. In particular, AP estimates the average of the precision values obtained across a range of different recall levels, ranging from $0$ to $1$. More specifically, AP integrates precision, recall, and the confidence scores associated with the model's detections, offering a measure of the model's ability to achieve high precision at various levels of recall. It is important to highlight that AP is calculated separately for each supported object class in a multi-class detection problem. Regarding its actual computation, AP measures the area under the Precision-Recall (PR) curve for a specific object class, where the PR curve is generated by varying the confidence threshold applied to the model's predictions, as follows:
\begin{equation}
  \text{AP}=\int_0^1{p(r)dr} \in [0,1],
\end{equation}
where $p(.)$ denotes the estimated PR curve.

Mean Average Precision (mAP) is an aggregated metric for evaluating the overall performance of object detection models, especially in scenarios involving multiple object classes. In particular, mAP is estimated by averaging the AP scores calculated for each individual object class, as follows:
\begin{equation}
  \text{mAP}=\frac{1}{N}\sum_n^N{\text{AP}_n},
\end{equation}
where $n$ denotes the index of each class and $N$ the total number of classes in the examined dataset. Among the different variants regarding how mAP is calculated, especially with respect to the defined IoU threshold for determining a true positive detection, the following ones have been considered in this work: a) \textbf{mAP$^\textbf{50}$}: This refers to the mAP score calculated using an IoU threshold of $0.5$. b) \textbf{mAP$^\textbf{50:95}$}: This involves a more rigorous evaluation protocol, which calculates mAP by averaging AP scores over a range of defined IoU thresholds, typically from $0.5$ to $0.95$ with a step of $0.05$, and subsequently averaging the computed results across all object classes. This metric provides a more comprehensive assessment of the model's localization accuracy, by considering its performance at different levels of overlap with the ground truth annotation.

\subsubsection{Time and computational complexity metrics}

In order to investigate the practical utility and widespread adoption of the considered object detectors, an analysis beyond solely their predictive accuracy is needed. Towards this direction, the following time and computational complexity metrics are considered in this study (the metrics are deeply interconnected, collectively dictating a model's efficiency, scalability, and suitability for real-world deployment):
\begin{itemize}

  \item \underline{Inference time}: It is typically measured in \textbf{milliseconds (ms)}, and it quantifies the duration a model requires to process a single input and to generate a corresponding prediction or decision. This metric fundamentally represents the time taken for a single forward propagation pass through the model's network architecture.

  \item \underline{Parameter size}: It is commonly expressed in \textbf{Millions (M)}, and it refers to the total count of trainable weights and biases present within a neural network model. These parameters are the fundamental adjustable values that the network learns and optimizes during its training phase.

  \item \underline{Computational load}: It is typically measured in \textbf{`Giga Floating-Point Operations per Second' (GFLOPS)} and it quantifies the computational performance of a model. Specifically, it represents the number of floating-point operations that can be performed per second, expressed in billions (giga).

\end{itemize}

\subsection{Implementation details}
\label{ssec:EF-implementation}

Regarding the implementation details of the object detectors described in Section \ref{ssec:EF-detectors}, the CNN backbones employed in this study, namely CSPDarkNet53 and HGNetV2, were pre-trained on the COCO dataset (implementation available in the Ultralytics\footnote{\url{https://github.com/ultralytics/ultralytics}} public toolbox). On the other hand, the transformer Swin-B backbone, pre-trained on ImageNet-22k, closely mirrors the configurations utilized by detectors such as DINO and Co-DETR. Additionally, the hybrid Next-ViT-S backbone was pre-trained on ImageNet-1k (weights provided by the authors\footnote{\url{https://github.com/bytedance/next-vit}}).

\begin{table}[!htbp]
  \centering
  \scriptsize
  \caption{Implementation details of the considered object detectors.}
  \begin{tabular}{
      m{4.8cm}
      m{4.8cm}
      m{1.2cm}
      m{1.0cm}
      m{1.0cm}
      m{1.0cm}
    }
    \toprule
    \textbf{Detection head}                                                                                                             & \textbf{Dataset}                                                                                                                                                                                & \textbf{Optimizer} & \textbf{Learning rate} & \textbf{Momentum} & \textbf{Weight decay} \\
    \midrule
    YOLOv8 \citep{jocherUltralyticsYOLOv82023}, RT-DETR \citep{lvDETRsBeatYOLOs2023}                                                    & OPIXray \citep{weiOccludedProhibitedItems2020}, CLCXray \citep{zhaoDetectingOverlappedObjects2022}, HiXray \citep{taoRealworldXraySecurity2021}, PIDray \citep{wangRealWorldProhibitedItem2021} & SGD                & 0.01                   & 0.9               & -                     \\
    \midrule
    YOLOv8 \citep{jocherUltralyticsYOLOv82023}, RT-DETR \citep{lvDETRsBeatYOLOs2023}                                                    & SIXray \citep{miaoSIXrayLargeScaleSecurity2019}, EDS \citep{taoExploringEndogenousShift2022}                                                                                                    & AdamW              & 0.000714               & 0.9               & 0.0006                \\
    \midrule
    CHR \citep{miaoSIXrayLargeScaleSecurity2019}, DOAM \citep{weiOccludedProhibitedItems2020}, LIM \citep{taoRealworldXraySecurity2021} & OPIXray \citep{weiOccludedProhibitedItems2020}, SIXray \citep{miaoSIXrayLargeScaleSecurity2019}                                                                                                 & AdamW              & 0.000714               & 0.9               & 0.0006                \\
    \midrule
    CHR \citep{miaoSIXrayLargeScaleSecurity2019}, DOAM \citep{weiOccludedProhibitedItems2020}, LIM \citep{taoRealworldXraySecurity2021} & CLCXray \citep{zhaoDetectingOverlappedObjects2022}, EDS \citep{taoExploringEndogenousShift2022}, PIDray \citep{wangRealWorldProhibitedItem2021}                                                 & SGD                & 0.01                   & 0.9               & -                     \\
    \midrule
    CHR \citep{miaoSIXrayLargeScaleSecurity2019}, LIM \citep{taoRealworldXraySecurity2021}                                              & HiXray \citep{taoRealworldXraySecurity2021}                                                                                                                                                     & AdamW              & 0.000714               & 0.9               & 0.0006                \\
    \midrule
    DOAM \citep{weiOccludedProhibitedItems2020}                                                                                         & HiXray \citep{taoRealworldXraySecurity2021}                                                                                                                                                     & SGD                & 0.01                   & 0.9               & -                     \\
    \midrule
    DINO \citep{zhangDINODETRImproved2022}                                                                                              & All                                                                                                                                                                                             & AdamW              & 0.0001                 & 0.9               & 0.0001                \\
    \midrule
    Co-DETR \citep{zongDETRsCollaborativeHybrid2023}                                                                                    & All                                                                                                                                                                                             & AdamW              & 0.0002                 & 0.9               & 0.0001                \\
    \bottomrule
  \end{tabular}
  \label{tab:hyperparameters}
\end{table}

Concerning the fine-tuning process, this varied across the various detectors. In particular, YOLOv8, RT-DETR, HR, DOAM, and LIM were trained for $100$ epochs, whereas the DINO and Co-DETR detectors were trained for $36$ epochs.  All experiments used early stopping to mitigate overfitting. YOLOv8 was fine-tuned with varying optimizers and learning rates across different datasets: Stochastic Gradient Descent (SGD) with a learning rate of $0.01$ and momentum ($\beta_{1}$) of $0.9$ for OPIXray, CLCXray, HiXray, and PIDray, and AdamW with a learning rate of $0.000714$, $\beta_{1}$  of $0.9$, and weight decay of $0.0006$ for SIXray and EDS. For RT-DETR similar configurations with YOLOv8 were used. HR, DOAM, and LIM were trained using AdamW with a learning rate of $0.000714$, $\beta_{1}$  of $0.9$, and weight decay of $0.0006$ for OPIXray and SIXray, while SGD with a learning rate of $0.01$ and momentum of $0.9$ was applied for CLCXray, EDS, and PIDray. In HiXray, HR and LIM were trained using AdamW with a learning rate of $0.000714$, momentum of $0.9$, and weight decay of $0.0006$, with DOAM being trained using SGD with a learning rate of $0.01$ and momentum of $0.9$. The DINO detector was trained on all datasets using its default settings, employing the AdamW optimizer with a learning rate of $0.0001$, $\beta_{1}$ set to $0.9$, and a weight decay of $0.0001$. Similarly, Co-DETR was trained on all datasets with its default configuration, utilizing the AdamW optimizer with a learning rate of $0.0002$, $\beta_{1}$ at $0.9$, and a weight decay of $0.0001$. Table \ref{tab:hyperparameters} provides a compact and comprehensive summary of the implementation details for the various object detectors considered in this work.

\section{Experimental results and insights}
\label{sec:EvaluationResults}

This section demonstrates the comparative evaluation results of the various DL-based X-ray object detection methods considered, according to the framework defined in Section \ref{sec:EvaluationFramework}, as well as critical observations and detailed insights. In order to thoroughly and systematically present the outcomes, the analysis is organized according to the following main axes/perspectives:
\begin{itemize}
  \item Overall performance of object detectors;
  \item Object-level detection results;
  \item Dataset-specific observations;
  \item Time efficiency and computational complexity aspects.
\end{itemize}

\subsection{Overall performance of object detectors}
\label{ssec:OverallInsights}

This section discusses the overall behavior of the various object detectors (Section \ref{ssec:EF-detectors}) considered in this work, across six of the most recent, large-scale and widely used public benchmarks (Section \ref{ssec:EF-datasets}). In particular, Table \ref{tab:results} illustrates the achieved overall object detection performance (mAP\textsuperscript{50} and mAP\textsuperscript{50:95} metrics reported) of the various detectors for the OPIXray \citep{weiOccludedProhibitedItems2020}, CLCXray \citep{zhaoDetectingOverlappedObjects2022}, SIXray \citep{miaoSIXrayLargeScaleSecurity2019}, EDS \citep{taoExploringEndogenousShift2022}, HiXray \citep{taoRealworldXraySecurity2021}, and PIDray \citep{wangRealWorldProhibitedItem2021} datasets. From the reported results, the following key observations can be made:

\begin{table*}[!htbp]
  \centering
  \footnotesize
  \caption{Object detection performance (mAP\textsuperscript{50}/mAP\textsuperscript{50:95}) for the OPIXray, CLCXray, SIXray, EDS, HiXray, and  PIDray datasets.}
  \begin{tabular}{
    m{3cm}
    >{\centering\arraybackslash}m{1.45cm}
    >{\centering\arraybackslash}m{1.45cm}
    >{\centering\arraybackslash}m{1.45cm}
    >{\centering\arraybackslash}m{1.45cm}
    >{\centering\arraybackslash}m{1.45cm}
    >{\centering\arraybackslash}m{1.45cm}
    c
    }
    \toprule
    \multirow{2.5}{*}{\textbf{Configuration}} & \multicolumn{6}{c}{\textbf{Dataset}} & \multirow{2.5}{*}{\textbf{Average}}                                                                                                                                 \\
    \cmidrule(lr){2-7}
                                              & \textbf{OPIXray}                     & \textbf{CLCXray}                    & \textbf{SIXray}        & \textbf{EDS (avg.)}    & \textbf{HIXray}        & \textbf{PIDray (overall)} &                        \\
    \midrule
    \multicolumn{6}{l}{\textbf{Generic CNN detectors}}                                                                                                                                                                                                     \\
    \midrule
    D(YOLOv8, CSPDarkNet53)                   & 0.868 / 0.413                        & 0.733 / 0.636                       & 0.901 / \textbf{0.794} & 0.547 / 0.386          & 0.845 / \textbf{0.564} & 0.897 / \textbf{0.807}    & 0.799 / 0.600          \\
    D(YOLOv8, HGNetV2)                        & 0.898 / 0.418                        & 0.725 / 0.617                       & 0.897 / 0.775          & 0.550 / 0.378          & 0.833 / 0.557          & 0.902 / 0.796             & 0.801 / 0.590          \\
    \midrule
    \multicolumn{6}{l}{\textbf{Custom CNN detectors}}                                                                                                                                                                                                      \\
    \midrule
    D(YOLOv8+CHR, CSPDarkNet53)               & 0.835 / 0.368                        & 0.710 / 0.602                       & 0.850 / 0.700          & 0.416 / 0.276          & 0.811 / 0.523          & 0.782 / 0.644             & 0.734 / 0.519          \\
    D(YOLOv8+DOAM, CSPDarkNet53)              & 0.790 / 0.361                        & 0.720 / 0.614                       & 0.828 / 0.658          & 0.422 / 0.280          & 0.830 / 0.545          & 0.815 / 0.689             & 0.734 / 0.525          \\
    D(YOLOv8+LIM, CSPDarkNet53)               & 0.791 / 0.344                        & 0.717 / 0.605                       & 0.827 / 0.661          & 0.446 / 0.300          & 0.828 / 0.525          & 0.800 / 0.664             & 0.735 / 0.517          \\
    \midrule
    \multicolumn{6}{l}{\textbf{Generic transformer detectors}}                                                                                                                                                                                             \\
    \midrule
    D(DINO, Swin-B)                           & 0.928 / 0.413                        & 0.739 / 0.607                       & 0.902 / 0.765          & 0.560 / 0.378          & 0.849 / 0.535          & 0.802 / 0.655             & 0.797 / 0.559          \\
    D(Co-DETR, Swin-B)                        & 0.928 / 0.423                        & 0.772  / \textbf{0.654}             & 0.893 / 0.735          & 0.653 / \textbf{0.450} & 0.857 / 0.531          & 0.852 / 0.732             & 0.826 / 0.587          \\
    \midrule
    \multicolumn{6}{l}{\textbf{Generic hybrid detectors}}                                                                                                                                                                                                  \\
    \midrule
    D(RT-DETR, HGNetV2)                       & 0.898 / 0.389                        & 0.721 / 0.609                       & 0.901 / 0.789          & 0.573 / 0.410          & 0.839 / 0.510          & 0.835 / 0.720             & 0.795 / 0.571          \\
    D(YOLOv8, Next-ViT-S)                     & 0.906 / \textbf{0.429}               & 0.740 / 0.640                       & 0.906 / 0.793          & 0.588 / 0.408          & 0.841 / 0.551          & 0.898 / 0.801             & 0.813 / \textbf{0.604} \\
    D(RT-DETR, Next-ViT-S)                    & 0.887 / 0.389                        & 0.720 / 0.609                       & 0.889 / 0.762          & 0.504 / 0.322          & 0.818 / 0.483          & 0.879 / 0.773             & 0.783 / 0.556          \\
    \bottomrule
  \end{tabular}
  \label{tab:results}
\end{table*}

\begin{itemize}
  \item \underline{General remarks}: There is \textbf{no single type of detector or class of methods (i.e, CNN, transformer, or hybrid) that is clearly shown advantageous} across all benchmarks. This critically highlights the need for an in depth performance analysis under multiple experimental settings, as this study does.

  \item \underline{Behavior of generic CNN detectors}: \textbf{CNN detectors exhibit the most consistent performance} for the considered architectural configurations across all benchmarks. A more careful analysis though reveals that \textbf{CNNs tend to be advantageous in relatively less challenging datasets} (e.g., PIDray and SIXray), but \textbf{their performance is inferior in more complex ones} (e.g., EDS (where domain distribution shifts are present, as will be detailed in Section \ref{ssec:DatasetInsights})). In particular, the D(YOLOv8, CSPDarkNet53) detector achieves the highest recognition rates in $3$ out of the $6$ considered benchmarks. This observed behavior is mainly due to the increased efficiency of the convolutional operators in modeling and recognizing local image patterns and correlations.

  \item \underline{Behavior of custom CNN detectors}: A counter-intuitive, but critical finding, of this study is the \textbf{consistent under-performance of custom CNN detectors} that incorporate X-ray-specific auxiliary modules. In particular, detectors D(YOLOv8+CHR, CSPDarkNet53), D(YOLOv8+DOAM, CSPDarkNet53) and D(YOLOv8+LIM, CSPDarkNet53) fall behind (and in most cases significantly) the generic D(YOLOv8, CSPDarkNet53) baseline across all considered datasets. The latter challenges the assumption that domain-specific modules (like CHR \citep{miaoSIXrayLargeScaleSecurity2019}, DOAM \citep{weiOccludedProhibitedItems2020}, and LIM  \citep{taoRealworldXraySecurity2021}) can always reinforce a state-of-art detector. More specifically, the computed results suggest that the integration of X-ray-specific auxiliary modules in modern CNN architectures like YOLOv8 (CHR, DOAM and LIM have been originally evaluated using earlier versions of the YOLO detection scheme) appears to lead to architectural disharmony and, thus, inferior performance.

  \item \underline{Behavior of generic transformer detectors}: \textbf{Transformer detectors demonstrate variations in performance} with respect to the selected architectural configuration. However, the D(Co-DETR, Swin-B) detector exhibits competitive performance to the one achieved by the CNN ones. In particular, D(Co-DETR, Swin-B) accomplishes the highest recognition rates in $2$ out of the $6$ considered benchmarks; interestingly, when compared only with D(YOLOv8, CSPDarkNet53), D(Co-DETR, Swin-B) is superior in half of the datasets. Notably, \textbf{D(Co-DETR, Swin-B) demonstrates increased performance in the most challenging benchmarks}; specifically, in the EDS dataset (presence of domain distribution shifts, as will be detailed in Section \ref{ssec:DatasetInsights}), it outperforms all other detectors. The latter suggests that the increased capability of the transformer blocks in modeling global context and long-range dependencies is beneficial in X-ray images that contain significantly cluttered scenes and variations in the data distribution (originating, for example, from the use of different X-ray inspection equipment).

  \item \underline{Behavior of hybrid CNN-transformer detectors}: Similarly to the case of transformers, \textbf{hybrid detection schemes demonstrate significant variations in performance} with respect to the selected architectural configuration. Interestingly, though, \textbf{the D(YOLOv8, Next-ViT-S) detector exhibits the best overall performance on average}. Additionally, \textbf{D(YOLOv8, Next-ViT-S) outperforms all other methods in the most challenging dataset} in this study, namely OPIXray (where significant object occlusions are present, as will be detailed in Section \ref{ssec:DatasetInsights}). This advantageous behavior of D(YOLOv8, Next-ViT-S) is mainly due to its hybrid architectural design, which relies on the combination of both convolutional (for modeling local image patterns) and transformer (for modeling global context and long-range dependencies) blocks.

  \item \underline{Effect of dataset size}: For the given benchmark scales, \textbf{the dataset size is not shown to exhibit a clear correlation with the performance achieved by the various detectors}. In particular, the dataset nature (i.e., complexity of cluttered scenes, degree of object occlusions, etc.) appears to have greater impact on the detection performance, compared to the total number of available images (OPIXray: 8,885, CLCXray: 9,565, SIXray: 11,401, EDS: 14,219, HIXray: 45,364, PIDray: 47,677).

  \item \underline{Effect of object number}: For the considered datasets, \textbf{the number of supported objects is also not shown to have a clear correlation with the performance achieved by the various detectors}. Again, the dataset nature has greater importance for the detection process, compared to the total number of available object types (OPIXray: 5, CLCXray: 12, SIXray: 6, EDS: 10, HIXray: 8, PIDray: 12).
\end{itemize}

\begin{table}[t]
  \centering
  \small
  \caption{Object size distribution across datasets.}
  \begin{tabular}{m{3.6cm} c c c c}
    \toprule
    \textbf{Dataset}                                   & \textbf{Total} & \textbf{Small} & \textbf{Medium} & \textbf{Large} \\
    \midrule
    OPIXray \citep{weiOccludedProhibitedItems2020}     & 1772           & -              & 1772            & -              \\
    \midrule
    CLCXray \citep{zhaoDetectingOverlappedObjects2022} & 1421           & 3              & 225             & 1193           \\
    \midrule
    SIXray \citep{miaoSIXrayLargeScaleSecurity2019}    & 2409           & 10             & 1027            & 1375           \\
    \midrule
    EDS \citep{taoExploringEndogenousShift2022}        & 31655          & 694            & 14174           & 16795          \\
    \hspace{2mm}$\mathcal{D}_{1}$                      & 11652          & 32             & 3139            & 8481           \\
    \hspace{2mm}$\mathcal{D}_{2}$                      & 10001          & 501            & 6565            & 2940           \\
    \hspace{2mm}$\mathcal{D}_{3}$                      & 10002          & 161            & 4470            & 5374           \\
    \midrule
    HiXray \citep{taoRealworldXraySecurity2021}        & 20476          & 3              & 2059            & 18415          \\
    \midrule
    PIDray \citep{wangRealWorldProhibitedItem2021}     & 23382          & 23382          & -               & -              \\
    \hspace{2mm}\textit{easy}                          & 9482           & 9482           & -               & -              \\
    \hspace{2mm}\textit{hard}                          & 8892           & 8892           & -               & -              \\
    \hspace{2mm}\textit{hidden}                        & 5008           & 5008           & -               & -              \\
    \bottomrule
  \end{tabular}
  \label{tab:objects-per-dataset}
\end{table}

\subsection{Object-level detection results}
\label{ssec:ObjectInsights}

\begin{figure*}[!t]
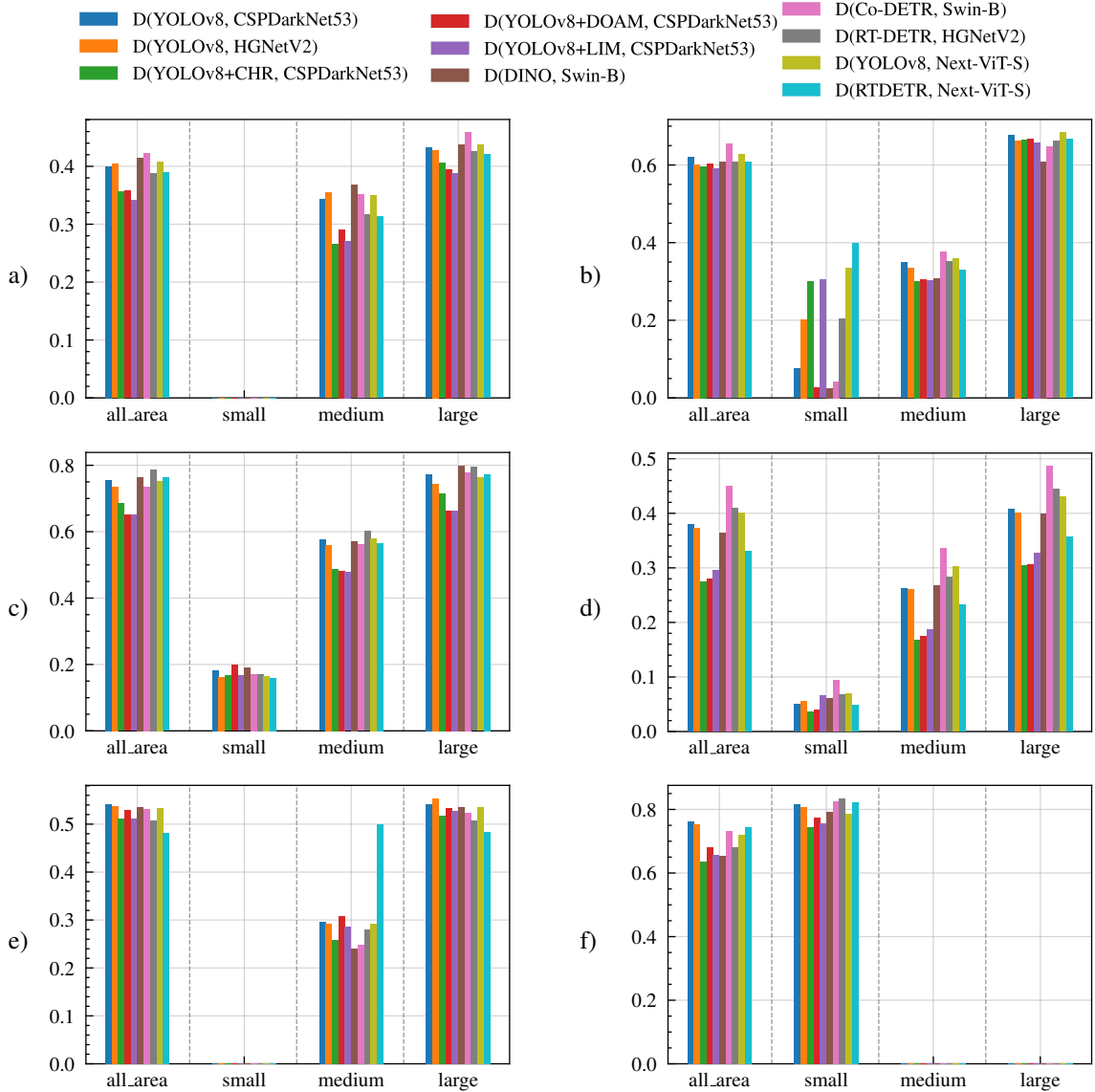

  \centering
  \begin{tabular}{p{0.001\linewidth}c  p{0.01\linewidth}c}
    \multicolumn{4}{c}{
      \begin{tabular}{c}
        \begin{subfigure}[t]{0.8\textwidth}
          \centering
          \includegraphics[width=\linewidth]{./Figures/legend.jpeg}
        \end{subfigure}
      \end{tabular}
    }                                   \\

    \begin{tabular}{c}a)\end{tabular} &

    \begin{tabular}{c}
      \begin{subfigure}[t]{0.4\textwidth}
        \centering
        \resizebox{\linewidth}{!}{\input{./Figures/Object_Size/OPIXray_mAP50-95_object-size.pgf}}
      \end{subfigure}
    \end{tabular}    &

    \begin{tabular}{c}b)\end{tabular} &

    \begin{tabular}{c}
      \begin{subfigure}[t]{0.4\textwidth}
        \centering
        \resizebox{\linewidth}{!}{\input{./Figures/Object_Size/CLCXray_mAP50-95_object-size.pgf}}
      \end{subfigure}
    \end{tabular}       \\

    \begin{tabular}{c}c)\end{tabular} &

    \begin{tabular}{c}
      \begin{subfigure}[t]{0.4\textwidth}
        \centering
        \resizebox{\linewidth}{!}{\input{./Figures/Object_Size/SIXray_mAP50-95_object-size.pgf}}
      \end{subfigure}
    \end{tabular}    &

    \begin{tabular}{c}d)\end{tabular} &

    \begin{tabular}{c}
      \begin{subfigure}[t]{0.4\textwidth}
        \centering
        \resizebox{\linewidth}{!}{\input{./Figures/Object_Size/EDS_avg_mAP50-95_object-size.pgf}}
      \end{subfigure}
    \end{tabular}       \\

    \begin{tabular}{c}e)\end{tabular} &

    \begin{tabular}{c}
      \begin{subfigure}[t]{0.4\textwidth}
        \centering
        \resizebox{\linewidth}{!}{\input{./Figures/Object_Size/HiXray_mAP50-95_object-size.pgf}}
      \end{subfigure}
    \end{tabular}    &

    \begin{tabular}{c}f)\end{tabular} &

    \begin{tabular}{c}
      \begin{subfigure}[t]{0.4\textwidth}
        \centering
        \resizebox{\linewidth}{!}{\input{./Figures/Object_Size/PIDray_avg_mAP50-95_object-size.pgf}}
      \end{subfigure}
    \end{tabular}       \\
  \end{tabular}

  \caption{Object-size detection performance (mAP$^{50:95}$ metric) for datasets: a) OPIXray, b) CLCXray, c) SIXray, d) EDS (avg.), e) HiXray, and f) PIDray (overall).}
  \label{fig:object-size-figures}
\end{figure*}

This section provides a systematic and granular analysis of object-level performance, in order to facilitate the generation of insights at a finer level of detail and a deeper understanding of the behavior of each detector. For that purpose, the object-level performance (only the mAP\textsuperscript{50:95} metric is provided) for all detectors and datasets considered in this work is illustrated in Fig. \ref{fig:object-level-figures}. Additionally, an object-size performance analysis is also implemented. In particular, according to the COCO \citep{linMicrosoftCOCOCommon2014} object detection dataset specifications, an object is classified as small if its area is less than $32^2$ pixels, medium if its area is between $32^2$ and $96^2$ pixels, and large if its area exceeds $96^2$ pixels. By adopting the aforementioned COCO definitions, Table \ref{tab:objects-per-dataset} summarizes the object size distribution for the datasets considered in this study, while Fig. \ref{fig:object-size-figures} depicts the corresponding object-size performance (only the mAP\textsuperscript{50:95} metric is provided). From the reported results, the following key observations can be made:

\begin{figure*}[!t]
  \centering
  \begin{tabular}{p{0.001\linewidth}c  p{0.01\linewidth}c}
    \multicolumn{4}{c}{
      \begin{tabular}{c}
        \begin{subfigure}[t]{0.8\textwidth}
          \centering
          \includegraphics[width=\linewidth]{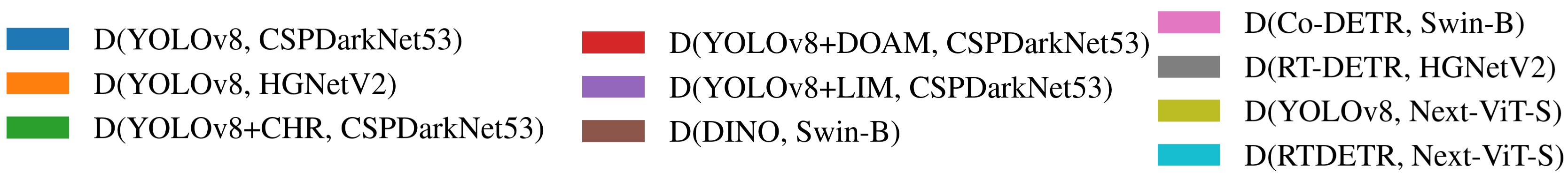}
        \end{subfigure}
      \end{tabular}
    }                                   \\

    \begin{tabular}{c}a)\end{tabular} &

    \begin{tabular}{c}
      \begin{subfigure}[t]{0.4\textwidth}
        \centering
        \resizebox{\linewidth}{!}{\input{./Figures/Class/OPIXray_mAP50-95_classwise.pgf}}
      \end{subfigure}
    \end{tabular}    &

    \begin{tabular}{c}b)\end{tabular} &

    \begin{tabular}{c}
      \begin{subfigure}[t]{0.4\textwidth}
        \centering
        \resizebox{\linewidth}{!}{\input{./Figures/Class/CLCXray_mAP50-95_classwise.pgf}}
      \end{subfigure}
    \end{tabular}       \\

    \begin{tabular}{c}c)\end{tabular} &

    \begin{tabular}{c}
      \begin{subfigure}[t]{0.4\textwidth}
        \centering
        \resizebox{\linewidth}{!}{\input{./Figures/Class/SIXray_mAP50-95_classwise.pgf}}
      \end{subfigure}
    \end{tabular}    &

    \begin{tabular}{c}d)\end{tabular} &

    \begin{tabular}{c}
      \begin{subfigure}[t]{0.4\textwidth}
        \centering
        \resizebox{\linewidth}{!}{\input{./Figures/Class/EDS_avg._mAP50-95_classwise.pgf}}
      \end{subfigure}
    \end{tabular}       \\

    \begin{tabular}{c}e)\end{tabular} &

    \begin{tabular}{c}
      \begin{subfigure}[t]{0.4\textwidth}
        \centering
        \resizebox{\linewidth}{!}{\input{./Figures/Class/HiXray_mAP50-95_classwise.pgf}}
      \end{subfigure}
    \end{tabular}    &

    \begin{tabular}{c}f)\end{tabular} &

    \begin{tabular}{c}
      \begin{subfigure}[t]{0.4\textwidth}
        \centering
        \resizebox{\linewidth}{!}{\input{./Figures/Class/PIDray_overall_mAP50-95_classwise.pgf}}
      \end{subfigure}
    \end{tabular}       \\
  \end{tabular}

  \caption{Object-level detection performance (mAP$^{50:95}$ metric) for datasets: a) OPIXray, b) CLCXray, c) SIXray, d) EDS (avg.), e) HiXray, and f) PIDray (overall).}
  \label{fig:object-level-figures}
\end{figure*}

\begin{itemize}
  \item \underline{General remarks}: \textbf{The physical properties of the objects at hand (e.g., material density, geometric complexity, size, etc.) influence heavily their detection performance}, regardless of the object detector considered (Fig. \ref{fig:object-size-figures}). In particular, object types that exhibit a relatively distinctive X-ray signature are: Scissor (SC) in OPIXray, Knife (KN) in CLCXray, Gun (GU) in SIXray, Laptop (LA) in EDS, Laptop (LA) in HiXray, and Wrench (WR) in PIDray; these mainly correspond to high-density metallic objects. On the contrary, objects types that are more difficult to discriminate are: Straight knife (ST) in OPIXray, Glass bottle (GB) in CLCXray, Wrench (WR) in SIXray, Knife (KN) in EDS, Nonmetallic Lighter (NL) in HiXray, and Gun (GU) in PIDray; these mainly constitute either low-density objects or objects with complex geometries. It needs to be highlighted though that the complexity of an object's X-ray signature constitutes a multi-factorial problem that arises from the interplay between the object's intrinsic properties (i.e., physical form and material composition) and its contextual presentation within the imaging system (i.e., object orientation and surrounding environment) \citep{mery2017logarithmic,viriyasaranon2022mfa}.

  \item \underline{Effect of material density}: The visual representation of an object in an X-ray image (i.e., its level of bright or dark appearance) is governed by a property known as the `linear attenuation coefficient' \citep{bai2003generalized}, which quantifies the fraction of the X-ray photons that are removed from a beam (either absorbed or scattered) as it passes through a unit thickness of a material. In this respect, an object appearing `dense'/`dark' on a scanner screen is one with a high such coefficient. While physical density (mass per unit volume) is a contributing factor, it is not the dominant one; the attenuation coefficient is a function of two fundamental material properties (namely, physical density and effective atomic number $Z_{eff}$) and the energy of the X-ray beam (i.e., photon energy) itself \citep{qi2010quantitative}. Taking into account the above analysis (while assuming for each object type its typical material composition and how its constituent materials interact with X-rays), the objects considered in this study can be roughly categorized into low- and high-complexity signatures \citep{bai2003generalized, qi2010quantitative}, as illustrated in Table \ref{tab:material-density}. From the results presented in Fig. \ref{fig:object-level-figures}, it can be seen that \textbf{objects with high density/attenuation exhibit higher detection rates} (e.g., knife, scissors, etc.), regardless of their signature complexity. On the contrary, \textbf{items with low density/attenuation are associated with lower recognition rates} (e.g., liquid in plastic bottle, carton drink, etc.). This performance difference is mainly grounded on the physics of X-ray imaging, where dense objects produce strong high-contrast signals that are distinguished more easily from the background.

        \begin{table}[t]
          \centering
          \footnotesize
          \caption{X-ray object signature classes grouped according to material complexity.}
          \label{tab:material-density}
          \renewcommand\tabularxcolumn[1]{m{#1}}
          \begin{tabular}{%
            >{\raggedright\arraybackslash}m{2cm}  
            >{\raggedright\arraybackslash}m{1.5cm}  
            >{\raggedright\arraybackslash}m{10cm}
            }                                      
            \toprule
            \textbf{Complexity level} & \textbf{Category}      & \textbf{Objects}                                                                                                                              \\
            \midrule
            \multirow{3.25}{=}{\textbf{Low} (homogeneous)} %
                                      & High-attenuation       & Gun (GU), Bullet (BU), Knife (KN, FO, ST, BL, DA), Wrench (WR), Pliers (PL), Hammer (HA), Handcuffs (HC), Baton (BA), Metal cans (CA, SP, TI) \\ \cmidrule{2-3}
                                      & Low-attenuation        & Water (WA), Liquid in plastic bottle (PL, DB), Non-metal lighter (NL), Cosmetic (CO), Carton drink (CD), Glass bottle (GB)                    \\
            \midrule
            \multirow{5.5}{=}{\textbf{High} (composite / electronic)} %
                                      & Mechanical composites  & Scissors (SC), Multi-tool knife (MU), Utility knife (UT), Swiss-army knife (SW), Lighter (LI), Umbrella (UM)                                  \\ \cmidrule{2-3}
                                      & Electronic devices     & Laptop (LA), Tablet (TA), Mobile phone (MP), Power bank (PB), Chargers (PO1, PO2), Device (SE)                                                \\ \cmidrule{2-3}
                                      & Pressurised containers & Vacuum cup (VA), Spray can (SP), Pressure vessel (PR)                                                                                         \\
            \bottomrule
          \end{tabular}
        \end{table}

  \item \underline{Effect of geometric complexity}: In the context of X-ray security imaging, the notion of the geometric complexity of an object comprises a multi-dimensional issue that goes significantly beyond the `simple' measurement of an object's shape. Typical factors that affect the object's visual appearance include, among others, the presence of heavy occlusion and overlapping, clutter, viewpoint dependency and geometric variations, and intra-class variance \citep{velayudhan2025sting, mery2017logarithmic}. Relying on this analysis \citep{rogers2017automated, liuBGMBackgroundMixup2024} (while also, importantly, considering common security screening practices and empirical operational evidence), the objects considered in this study can be roughly graded in terms of their exhibited geometric complexity, taking into account the following key aspects: structural complexity, material and density profile, and contextual complexity; the resulting geometric complexity classification is shown in Table \ref{tab:material-geometry}. From the results presented in Fig. \ref{fig:object-level-figures}, it can be seen that \textbf{objects with low or moderate geometric complexity tend to be associated with increased detection rates} (e.g., dagger, wrench, etc.). On the contrary, \textbf{items with high or very high complexity have a tendency towards decreased performance} (e.g., pressure, gun, etc.).

        \begin{table}[t]
          \centering
          \footnotesize
          \caption{X-ray object classes grouped according to geometric complexity.}
          \label{tab:material-geometry}
          \renewcommand\tabularxcolumn[1]{m{#1}}

          \renewcommand{\arraystretch}{1.1}
          \begin{tabular}{%
            >{\raggedright\arraybackslash}m{1.3cm}  
            >{\raggedright\arraybackslash}m{5cm}  
            >{\raggedright\arraybackslash}m{8cm}
            }                                      
            \toprule
            \textbf{Complexity level} & \textbf{Description}                                                                                        & \textbf{Objects}                                                                                                                                                               \\
            \midrule
            \textbf{Low}              & Simple, monolithic shapes                                                                                   & Knife (KN), Blade (BL), Straight knife (ST), Dagger (DA), Bullet (BU), Water (WA), Carton drink (CD), Glass bottle (GB), Plastic bottle (PL, DB)                               \\
            \midrule
            \textbf{Moderate}         & Defined shapes with simple articulation or composite materials                                              & Wrench (WR), Pliers (PL), Hammer (HA), Handcuffs (HC), Baton (BA), Scissors (SC), Can (CA, SP), Tin (TI), Lighter (LI), Cosmetic (CO), Vacuum cup (VA), Non-metal lighter (NL) \\
            \midrule
            \textbf{High}             & Significant articulation and state-dependent variability                                                    & Folding knife (FO), Utility knife (UT), Swiss-army knife (SW), Multi-tool knife (MU), Umbrella (UM), Pressure vessel (PR)                                                      \\
            \midrule
            \textbf{Very high}        & Extreme internal component density, composite materials, high radiopacity, potential to obscure other items & Gun (GU), Power bank (PB), Portable chargers (PO1, PO2), Laptop (LA), Mobile phone (MP), Tablet (TA), Device (SE)                                                              \\
            \bottomrule
          \end{tabular}
        \end{table}

  \item \underline{Effect of object size}: From the results illustrated in Fig. \ref{fig:object-size-figures}, it can be observed that \textbf{larger objects exhibit significantly increased detection rates}. In particular, as the size of the depicted items increases (from small to large), the detection performance improves correspondingly for all detectors and across all datasets.

  \item \underline{Effect of detector architectural configuration}: From the results depicted in Fig. \ref{fig:object-level-figures} and Fig. \ref{fig:object-size-figures}, it is shown that different detector architectural configurations are favorable for different types of objects. In particular, \textbf{transformer and hybrid (with transformer backbone) detectors generally show increased performance for larger and/or uniformly shaped items} (e.g., Straight knife (ST) in OPIXray, Glass bottle (GB) in CLCXray, Pressure PR in EDS, Device (SE) in EDS, Power bank (PB) in EDS, Laptop (LA) in EDS, etc.). On the other hand, \textbf{CNN detectors are favorable for smaller and/or more variably shaped items} (e.g., Portable chargers (PO1, PO2) in HiXray, Knife (KN) in PIDray, Hammer (HA) in PIDray, Sprayer SP in PIDray, etc.). The latter is largely explained by the fact that the large/global receptive field of transformers is shown to be advantageous for identifying large and contiguous objects. On the other hand, the strong local feature extraction and spatial invariance of CNNs appear to be efficient for detecting small and changeable shape items.

\end{itemize}

\subsection{Dataset-specific observations}
\label{ssec:DatasetInsights}

Apart from the analysis regarding the overall behavior of the various
object detectors (Section \ref{ssec:OverallInsights}) and the object-level performance (Section \ref{ssec:ObjectInsights}), this section focuses on investigating how the nature and the individual particularities/challenges of each dataset (e.g., dataset creation/capturing process, degree of object occlusions, use of different scanning machinery (domain shift), degree of clutter, range of object sizes, etc.) affect the recognition performance. From the computed results, the following key insights can be extracted:

\begin{itemize}
  \item \underline{OPIXray}: This benchmark focuses on investigating the robustness of detection schemes under real-world inspection scenarios and varying degrees of object occlusions (Section \ref{ssec:EF-datasets}). It contains only medium-sized objects (Table \ref{tab:objects-per-dataset}), while the test set is split into the following subsets: a) OL1: No or slight object occlusion, OL2: Partial item occlusion, and OL3: Severely or full object occlusion. The detailed results for all test subsets, as well as overall, are depicted in Table \ref{tab:results-opixray}. The reported results illustrate the superiority of the hybrid D(YOLOv8, Next-ViT-S) detector for all subsets, followed in principle by other transformer architectural schemes. The latter suggests that \textbf{the combination of transformer blocks (for learning global context and long-range dependencies) with convolutional ones (for modeling local image patterns) comprises a robust solution for handling object occlusions (at different levels)}. Moreover, the performance of all detectors naturally drops as the degree of occlusion increases.

        \begin{table*}
          \centering
          \footnotesize
          \caption{Object detection performance (mAP\textsuperscript{50}/mAP\textsuperscript{50:95}) for all test subsets of the OPIXray dataset.}
          \begin{tabular}{l *{4}{c}}
            \toprule
            \multirow{2}{*}{\textbf{Configuration}} & \multicolumn{4}{c}{\textbf{Dataset}}                                                                            \\
            \cmidrule(lr){2-5}
                                                    & \textbf{OL1}                         & \textbf{OL2}           & \textbf{OL3}           & \textbf{Overall}       \\
            \midrule
            \multicolumn{5}{l}{\textbf{Generic CNN detectors}}                                                                                                        \\
            \midrule
            D(YOLOv8, CSPDarkNet53)                 & 0.877 / 0.425                        & 0.865 / 0.407          & 0.852 / 0.400          & 0.868 / 0.413          \\
            D(YOLOv8, HGNetV2)                      & 0.917 / 0.432                        & 0.890 / 0.421          & 0.873 / 0.388          & 0.898 / 0.418          \\
            \midrule
            \multicolumn{5}{l}{\textbf{Custom CNN detectors}}                                                                                                         \\
            \midrule
            D(YOLOv8+CHR, CSPDarkNet53)             & 0.851 / 0.385                        & 0.839 / 0.362          & 0.802 / 0.354          & 0.835 / 0.368          \\
            D(YOLOv8+DOAM, CSPDarkNet53)            & 0.794 / 0.377                        & 0.810 / 0.361          & 0.773 / 0.344          & 0.790 / 0.361          \\
            D(YOLOv8+LIM, CSPDarkNet53)             & 0.806 / 0.356                        & 0.792 / 0.344          & 0.765 / 0.335          & 0.791 / 0.344          \\
            \midrule
            \multicolumn{5}{l}{\textbf{Generic transformer detectors}}                                                                                                \\
            \midrule
            D(DINO, Swin-B)                         & 0.934 / 0.430                        & 0.932 / 0.412          & 0.905 / 0.388          & 0.928 / 0.413          \\
            D(Co-DETR, Swin-B)                      & 0.933 / 0.437                        & 0.922 / 0.414          & 0.921 / 0.415          & 0.928 / 0.423          \\
            \midrule
            \multicolumn{5}{l}{\textbf{Generic hybrid detectors}}                                                                                                     \\
            \midrule
            D(RT-DETR, HGNetV2)                     & 0.905 / 0.402                        & 0.908 / 0.392          & 0.857 / 0.368          & 0.898 / 0.389          \\
            D(YOLOv8, Next-ViT-S)                   & 0.919 / \textbf{0.444}               & 0.896 / \textbf{0.424} & 0.894 / \textbf{0.417} & 0.906 / \textbf{0.429} \\
            D(RT-DETR, Next-ViT-S)                  & 0.904 / 0.400                        & 0.865 / 0.389          & 0.883 / 0.371          & 0.887 / 0.389          \\
            \bottomrule
          \end{tabular}
          \label{tab:results-opixray}
        \end{table*}

  \item \underline{CLCXray}: This dataset pays particular attention on examining object overlaps with same-class instances as well as with their surrounding background (Section \ref{ssec:EF-datasets}). It predominantly contains large objects (Table \ref{tab:objects-per-dataset}) and also liquid containers (apart from other common prohibited items). The reported results (Table \ref{tab:results}) indicate that \textbf{transformer and hybrid (with transformer backbone network) detectors outperform CNN ones}, mainly due to the efficiency of transformer blocks in recognizing large objects (as also explained in Section \ref{ssec:ObjectInsights}).

  \item \underline{SIXray}: This benchmark investigates real-world inspection scenarios (Section \ref{ssec:EF-datasets}), including medium/large objects (Table \ref{tab:objects-per-dataset}), as well as items with significant intra-class shape diversity. According to the performed experiments (Table \ref{tab:results}), \textbf{CNN and hybrid (with convolutional detection head) perform better}, in principle due to the efficiency of convolutional units in modeling variably shaped items (as also detailed in Section \ref{ssec:ObjectInsights}).

  \item \underline{EDS}: This dataset emphasizes on examining the effect of domain shift, by employing three different X-ray scanners during the data collection phase (Section \ref{ssec:EF-datasets}), while containing in principle medium/large-sized objects (Table \ref{tab:objects-per-dataset}). According to the benchmark's experimental protocol, a detection model in trained on a single domain (out of the three available in total) and evaluated on a different one; the detailed results for all six experimental sessions performed are depicted in Table \ref{tab:results-eds}. The reported results demonstrate the clear superiority of the transformer D(Co-DETR, Swin-b) detector for all sessions, followed by other hybrid architectural schemes. The latter illustrates the \textbf{increased ability of the transformer architectural blocks to handle domain distribution shifts}. Additionally, it suggests that transformer blocks tend to learn more fundamental/abstract object representations (i.e., across different types of scanners), mainly due to the inherent capability of attention mechanisms to model broader/global contextual information.

        \begin{table*}
          \centering
          \footnotesize
          \caption{Object detection performance (mAP\textsuperscript{50}/mAP\textsuperscript{50:95}) for all experimental sessions of the EDS dataset. $\mathcal{D}_{m\rightarrow n}$ indicates training on the $m^{th}$ domain/scanner and evaluation on the $n^{th}$ one.}
          \begin{tabular}{m{2.5cm} *{7}{c}}
            \toprule
            \multirow{2}{*}{\textbf{Configuration}} & \multicolumn{7}{c}{\textbf{Domain}}                                                                                                                                                                                                                                          \\
            \cmidrule(lr){2-8}
                                                    & \boldmath$\mathcal{D}_{1\rightarrow2}$ & \boldmath$\mathcal{D}_{1\rightarrow3}$ & \boldmath$\mathcal{D}_{2\rightarrow1}$ & \boldmath$\mathcal{D}_{2\rightarrow3}$ & \boldmath$\mathcal{D}_{3\rightarrow1}$ & \boldmath$\mathcal{D}_{3\rightarrow2}$ &
            \textbf{Avg.}                                                                                                                                                                                                                                                                                                          \\
            \midrule
            \multicolumn{8}{l}{\textbf{Generic CNN methods}}                                                                                                                                                                                                                                                                       \\
            \midrule
            D(YOLOv8, CSPDarkNet53)                 & 0.482 / 0.340                          & 0.555 / 0.410                          & 0.454 / 0.295                          & 0.619 / 0.449                          & 0.587 / 0.411                          & 0.590 / 0.411                          & 0.547 / 0.386          \\
            D(YOLOv8, HGNetV2)                      & 0.479 / 0.323                          & 0.558 / 0.403                          & 0.493 / 0.312                          & 0.610 / 0.429                          & 0.574 / 0.403                          & 0.586 / 0.398                          & 0.550 / 0.378          \\
            \midrule
            \multicolumn{8}{l}{\textbf{Custom CNN methods}}                                                                                                                                                                                                                                                                        \\
            \midrule
            D(YOLOv8+CHR, CSPDarkNet53)             & 0.342 / 0.226                          & 0.435 / 0.304                          & 0.352 / 0.222                          & 0.471 / 0.319                          & 0.452 / 0.296                          & 0.422 / 0.291                          & 0.416 / 0.276          \\
            D(YOLOv8+DOAM, CSPDarkNet53)            & 0.380 / 0.252                          & 0.432 / 0.295                          & 0.363 / 0.228                          & 0.386 / 0.250                          & 0.479 / 0.322                          & 0.492 / 0.335                          & 0.422 / 0.280          \\
            D(YOLOv8+LIM, CSPDarkNet53)             & 0.350 / 0.231                          & 0.416 / 0.285                          & 0.383 / 0.243                          & 0.534 / 0.370                          & 0.496 / 0.340                          & 0.495 / 0.333                          & 0.446 / 0.300          \\
            \midrule
            \multicolumn{8}{l}{\textbf{Generic transformer methods}}                                                                                                                                                                                                                                                               \\
            \midrule
            D(DINO, Swin-b)                         & 0.404 / 0.270                          & 0.612 / 0.437                          & 0.452 / 0.292                          & 0.643 / 0.439                          & 0.607 / 0.407                          & 0.645 / 0.426                          & 0.560 / 0.378          \\
            D(Co-DETR, Swin-b)                      & 0.557 / \textbf{0.386}                 & 0.680 / \textbf{0.503}                 & 0.572 / \textbf{0.361}                 & 0.702 / \textbf{0.502}                 & 0.692 / \textbf{0.483}                 & 0.701 / \textbf{0.467}                 & 0.653 / \textbf{0.450} \\
            \midrule
            \multicolumn{8}{l}{\textbf{Generic hybrid methods}}                                                                                                                                                                                                                                                                    \\
            \midrule
            D(RT-DETR-l, HGNetV2)                   & 0.506 / 0.352                          & 0.569 / 0.424                          & 0.506 / 0.350                          & 0.648 / 0.471                          & 0.595 / 0.431                          & 0.616 / 0.429                          & 0.573 / 0.410          \\
            D(YOLOv8, Next-ViT-s)                   & 0.512 / 0.347                          & 0.603 / 0.441                          & 0.515 / 0.341                          & 0.648 / 0.454                          & 0.624 / 0.434                          & 0.626 / 0.431                          & 0.588 / 0.408          \\
            D(RT-DETR-l, Next-ViT-s)                & 0.446 / 0.292                          & 0.545 / 0.343                          & 0.372 / 0.217                          & 0.450 / 0.286                          & 0.578 / 0.377                          & 0.636 / 0.419                          & 0.504 / 0.322          \\
            \bottomrule
          \end{tabular}
          \label{tab:results-eds}
        \end{table*}

  \item \underline{HIXray}: This benchmark incorporates object types that of interest in airport inspection scenarios (Section \ref{ssec:EF-datasets}), including predominantly large- and to a smaller extent medium-sized objects (Table \ref{tab:objects-per-dataset}). According to the reported results (Table \ref{tab:results}), \textbf{CNN detectors are advantageous (followed in principle by hybrid ones)}, mainly due to the fact that the dataset contains items with relatively fine-grained X-ray signatures (i.e., objects comprising multiple smaller and of high variance parts, like portable chargers, mobile phones, etc.) that convolutional operators are better in modeling local image characteristics and invariances.

  \item \underline{PIDray}: This dataset focuses on the detection of deliberately hidden items (Section \ref{ssec:EF-datasets}), including only small-sized objects (Table \ref{tab:objects-per-dataset}). Additionally, the test set is split into the following subsets: a) Easy: Only one prohibited object, b) Hard: More than one illicit items, and c) Hidden: Deliberately hidden objects. The detailed results for all test subsets, as well as overall, are depicted in Table \ref{tab:results-pidray}. From the reported results, it can be seen that \textbf{CNN detectors perform better for most scenarios, followed by in principle hybrid detectors (with a convolutional detection head)}. The latter is mainly due to the increased ability of the convolutional filters to model small-scale image features and objects with large fine-grained intra-class variance. Moreover, the performance of all detectors naturally drops for the most challenging `hidden' scenario.

        \begin{table*}
          \centering
          \footnotesize
          \caption{Object detection performance (mAP\textsuperscript{50}/mAP\textsuperscript{50:95}) for all test subsets of the PIDray dataset.}
          \begin{tabular}{l *{4}{c}}
            \toprule
            \multirow{2}{*}{\textbf{Configuration}} & \multicolumn{4}{c}{\textbf{Dataset}}                                                                            \\
            \cmidrule(lr){2-5}
                                                    & \textbf{Easy}                        & \textbf{Hard}          & \textbf{Hidden}        & \textbf{Overall}       \\
            \midrule
            \multicolumn{5}{l}{\textbf{Generic CNN detectors}}                                                                                                        \\
            \midrule
            D(YOLOv8, CSPDarkNet53)                 & 0.911 / \textbf{0.846}               & 0.914 / \textbf{0.812} & 0.797 / 0.682          & 0.874 / \textbf{0.780} \\
            D(YOLOv8, HGNetV2)                      & 0.918 / 0.840                        & 0.918 / 0.796          & 0.804 / 0.666          & 0.880 / 0.767          \\
            \midrule
            \multicolumn{5}{l}{\textbf{Custom CNN detectors}}                                                                                                         \\
            \midrule
            D(YOLOv8+CHR, CSPDarkNet53)             & 0.832 / 0.734                        & 0.824 / 0.656          & 0.691 / 0.541          & 0.783 / 0.644          \\
            D(YOLOv8+DOAM, CSPDarkNet53)            & 0.870 / 0.774                        & 0.873 / 0.725          & 0.702 / 0.567          & 0.815 / 0.689          \\
            D(YOLOv8+LIM, CSPDarkNet53)             & 0.855 / 0.750                        & 0.866 / 0.709          & 0.678 / 0.533          & 0.800 / 0.664          \\
            \midrule
            \multicolumn{5}{l}{\textbf{Generic transformer detectors}}                                                                                                \\
            \midrule
            D(DINO, Swin-B)                         & 0.884 / 0.771                        & 0.838 / 0.655          & 0.684 / 0.538          & 0.802 / 0.655          \\
            D(Co-DETR, Swin-B)                      & 0.904 / 0.819                        & 0.911 / 0.770          & 0.741 / 0.607          & 0.852 / 0.732          \\
            \midrule
            \multicolumn{5}{l}{\textbf{Generic hybrid detectors}}                                                                                                     \\
            \midrule
            D(RT-DETR, HGNetV2)                     & 0.864 / 0.780                        & 0.864 / 0.724          & 0.681 / 0.548          & 0.803 / 0.684          \\
            D(YOLOv8, Next-ViT-S)                   & 0.912/ 0.837                         & 0.910 / 0.799          & 0.803 / \textbf{0.685} & 0.842 / 0.736          \\
            D(RT-DETR, Next-ViT-S)                  & 0.898 / 0.824                        & 0.898 / 0.770          & 0.779 / 0.646          & 0.858 / 0.746          \\
            \bottomrule
          \end{tabular}
          \label{tab:results-pidray}
        \end{table*}

\end{itemize}

\subsection{Time efficiency and computational complexity aspects}
\label{ssec:TimeInsights}

\begin{figure*}[t]
  \centering
  \scriptsize

  \begin{subfigure}[t]{0.45\textwidth}
    \centering
    \resizebox{\linewidth}{!}{\input{Figures/complexity.pgf}}
    \label{fig:complexity}
  \end{subfigure}%
  \hspace{0.05\textwidth}
  \begin{subfigure}[t]{0.3\textwidth}
    \centering
    \raisebox{3\baselineskip}{
      \includegraphics[width=\linewidth]{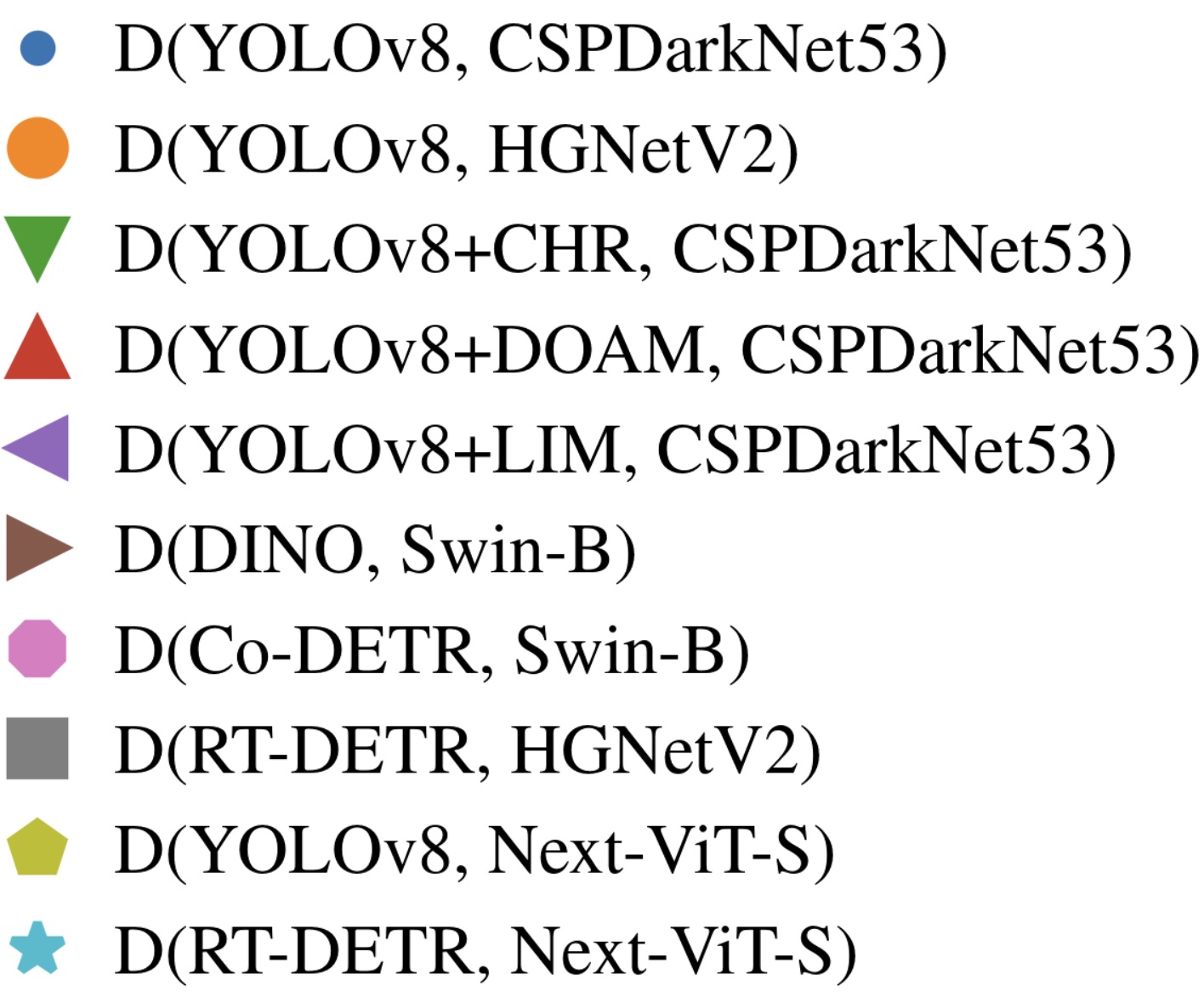}
    }
    \label{fig:complexity-legend}
  \end{subfigure}

  \caption{Object detector complexity analysis diagram (inference time (ms) vs. parameter size (M)).}
  \label{fig:complexity-overall}
\end{figure*}

Complementary to the analysis regarding the recognition performance of the various detectors (Sections \ref{ssec:OverallInsights}-\ref{ssec:DatasetInsights}), this section emphasizes on investigating time performance and computational complexity aspects, aiming at shedding light on practical issues concerning feasibility for real-world deployment. In particular, Table \ref{tab:complexity-evaluation} illustrates the estimated time efficiency and computational complexity metrics (namely, inference time\footnote{Measured on an NVIDIA RTX 4070 Ti GPU.} (ms), parameter size (M), and computational load (GFLOPS)) for the various detectors considered in this study. Additionally, a complexity analysis diagram (plotting inference time against parameter size per model/detector) is provided in Fig. \ref{fig:complexity-overall}, in order to better demonstrate the relation between the number of parameters and the actual time performance for each model. From the computed results, the following key insights can be extracted:

\begin{table}[t]
  \centering
  \footnotesize
  \caption{Object detector time efficiency and computational complexity analysis.}
  \begin{tabularx}{0.8\textwidth}{m{4.3cm} Y Y Y}
    \toprule
    \multirow{2}{*}{\textbf{Configuration}} & \multirow{2}{*}{\textbf{Inference time (ms)}} & \multirow{2}{*}{\textbf{Parameter size (M)}} & \textbf{Computational load (GFLOPS)} \\
    \midrule
    \multicolumn{4}{l}{\textbf{Generic CNN methods}}                                                                                                                              \\
    \midrule
    D(YOLOv8, CSPDarkNet53)                 & 7.52                                          & 43.6                                         & 165.4                                \\
    D(YOLOv8, HGNetV2)                      & \textbf{5.3}                                  & 38.5                                         & 128.6                                \\
    \midrule
    \multicolumn{4}{l}{\textbf{Custom CNN methods}}                                                                                                                               \\
    \midrule
    D(YOLOv8+CHR, CSPDarkNet53)             & 5.6                                           & 37.9                                         & \textbf{102.3}                       \\
    D(YOLOv8+DOAM, CSPDarkNet53)            & 27.1                                          & 43.6                                         & -                                    \\
    D(YOLOv8+LIM, CSPDarkNet53)             & 15.9                                          & 40.8                                         & 137.6                                \\
    \midrule
    \multicolumn{4}{l}{\textbf{Generic transformer methods}}                                                                                                                      \\
    \midrule
    D(DINO, Swin-B)                         & 159.85                                        & 108                                          & 560                                  \\
    D(Co-DETR, Swin-B)                      & 187                                           & 125                                          & 1068                                 \\
    \midrule
    \multicolumn{4}{l}{\textbf{Generic hybrid methods}}                                                                                                                           \\
    \midrule
    D(RT-DETR, HGNetV2)                     & 9.32                                          & \textbf{32}                                  & 110                                  \\
    D(YOLOv8, Next-ViT-S)                   & 12.22                                         & 56                                           & 174.9                                \\
    D(RT-DETR, Next-ViT-S)                  & 16.52                                         & 48.7                                         & 121                                  \\
    \bottomrule
  \end{tabularx}
  \label{tab:complexity-evaluation}
\end{table}

\begin{itemize}
  \item \underline{Time performance}: From the results presented in Table \ref{tab:complexity-evaluation}, it can be seen that the time efficiency of the various object detectors can be graded and roughly classified in the following main categories:
        \begin{itemize}
          \item High-throughput inference ($<$10ms): \textbf{Generic CNN detectors exhibit increased (real-time) inference speed}, making them suitable for demanding real-world application settings. The latter is mainly due to the careful design of the involved architectures, optimized so as to achieve high through-put rates.
          \item Moderate inference (10-20ms): \textbf{Hybrid architectures accomplish moderate inference rates}, where those that incorporate a CNN backbone perform faster. This decrease in time efficiency (compared to the CNN case) is twofold: a) The integration of both CNN and transformer blocks in the overall architecture; this combination is not optimized regarding inference speed, and b) When transformers are integrated in the overall model, these typically correspond to larger networks that inevitably lead to decreased inference rates.
          \item Laggy inference (150ms+): \textbf{Pure transformer detectors exhibit decreased processing speed}, which is mainly due to the significantly increased number of involved model parameters (compared to the other detector categories), as discussed above.
        \end{itemize}

  \item \underline{Architectural optimization}: Theoretical complexity metrics, like parameter number and GFLOPS, do not always correspond to accurate real-world latency estimations. For example, the D(YOLOv8+LIM, CSPDarkNet53) detector, although it exhibits a comparable parameter number with and a lower GFLOPS rate than D(YOLOv8, CSPDarkNet53), performs approximately two times slower than D(YOLOv8, CSPDarkNet53) (Table \ref{tab:complexity-evaluation} and Fig. \ref{fig:complexity-overall}). The latter clearly demonstrates that \textbf{the nature of the computational operations and their suitability for parallelization on GPU hardware is of paramount importance and significantly affects inference time}.

  \item \underline{Cost of customization}: Apart from degrading detection performance (Section \ref{ssec:OverallInsights}), \textbf{custom CNN architectures are shown to introduce significant burdens on inference speed in most cases} (Table \ref{tab:complexity-evaluation}). In particular, the D(YOLOv8+DOAM, CSPDarkNet53) and D(YOLOv8+LIM, CSPDarkNet53) detectors perform approximately four and two times slower than their baseline counterpart D(YOLOv8, CSPDarkNet53), respectively. This suggests that custom modules (namely, DOAM, and LIM), although not computationally demanding in theory, may introduce sequential operations or complex memory access patterns that are inefficient on a GPU accelerator; hence, leading to a corresponding latency bottleneck.

\end{itemize}

In order to summarize the key findings of the current work (as detailed in Sections \ref{ssec:OverallInsights}-\ref{ssec:TimeInsights}), the main insights derived from the performed comparative evaluation study are illustrated in Table \ref{tab:overallInsights}.

\begin{table*}[!ht]
  \small
  \caption{Main insights derived from the performed comparative evaluation study.}
  \label{tab:overallInsights}

  \begingroup
  \renewcommand{\arraystretch}{1.2}

  \begin{threeparttable}
    \footnotesize
    \renewcommand\tabularxcolumn[1]{m{#1}}
    \begin{tabularx}{\textwidth} {X X X X X}
      \toprule
      \multicolumn{5}{c}{\textbf{General insights}}                                                                                                                                             \\
      \midrule
      \multicolumn{5}{>{\raggedright\arraybackslash}p{\dimexpr\textwidth-2\tabcolsep\relax}}{
      \begin{itemize}[nosep,leftmargin=*]
        \item No single type of detector clearly advantageous across all settings
        \item No clear correlation of dataset size with detection performance
        \item No clear correlation of number of supported objects with detection performance
        \item The physical properties of the objects influence heavily their detection
        \item Increased detection rates for high density/attenuation objects
        \item Decreased detection rates for low density/attenuation items
        \item Improved detection performance for objects with low or moderate geometric complexity
        \item Tendency towards decreased detection performance for items with high or very high geometric complexity
        \item Significantly increased detection rates for large objects
        \item Inference time significantly affected by the nature of the computational operations and their suitability for parallelization on GPU hardware
      \end{itemize}
      }                                                                                                                                                                                         \\
      \midrule
      \multicolumn{5}{c}{\textbf{Detector type-specific insights}}                                                                                                                              \\
      \midrule
      \textbf{Aspect}                                                     & \textbf{Generic CNN} & \textbf{Custom CNN} & \textbf{Generic transformer} & \textbf{Generic hybrid CNN-transformer} \\
      \midrule
      \textbf{Overall performance}                                        &
      \begin{itemize}[nosep,leftmargin=*]
        \item Most consistent performance
        \item Advantageous in less challenging settings
        \item Inferior performance in more complex benchmarks
      \end{itemize}               &
      \begin{itemize}[nosep,leftmargin=*]
        \item Consistent under-performance
      \end{itemize}                                 &
      \begin{itemize}[nosep,leftmargin=*]
        \item Variations in performance for different detectors
        \item Increased performance in the most challenging benchmarks
      \end{itemize}      &
      \begin{itemize}[nosep,leftmargin=*]
        \item Significant variations in performance for different detectors
        \item Best overall performance on average
        \item Best performance in the most challenging dataset
      \end{itemize}                                                                                                                        \\
      \midrule
      \textbf{Object-level detection}                                     &
      \begin{itemize}[nosep,leftmargin=*]
        \item Favorable for small and/or variably shaped items
      \end{itemize}              &
      \begin{itemize}[nosep,leftmargin=*]
        \item Favorable for small and/or variably shaped items
      \end{itemize}              &
      \begin{itemize}[nosep,leftmargin=*]
        \item Increased performance for large and/or uniformly shaped items
      \end{itemize} &
      \begin{itemize}[nosep,leftmargin=*]
        \item Increased performance for large and/or uniformly shaped items (transformer backbone)
      \end{itemize}                                                                                                 \\
      \midrule
      \textbf{Dataset-specific observations}                              &
      \begin{itemize}[nosep,leftmargin=*]
        \item Perform best for most benchmarks
        \item Efficient for deliberately hidden items
      \end{itemize}                       &
      \begin{itemize}[nosep,leftmargin=*]
        \item --
      \end{itemize}                                 &
      \begin{itemize}[nosep,leftmargin=*]
        \item Robust to domain distribution shifts
      \end{itemize}                          &
      \begin{itemize}[nosep,leftmargin=*]
        \item Robust to object occlusions (at different levels)
        \item Robust to domain distribution shifts
      \end{itemize}                                                                                                                                    \\
      \midrule
      \textbf{Time efficiency}                                            &
      \begin{itemize}[nosep,leftmargin=*]
        \item High-throughput real-time inference ($<\!10$ ms)
      \end{itemize}              &
      \begin{itemize}[nosep,leftmargin=*]
        \item Significant burdens on inference speed in most cases
      \end{itemize}          &
      \begin{itemize}[nosep,leftmargin=*]
        \item Laggy inference (150 ms+)
      \end{itemize}                                 &
      \begin{itemize}[nosep,leftmargin=*]
        \item Moderate inference (10–20 ms)
      \end{itemize}                                                                                                                                                        \\
      \bottomrule
    \end{tabularx}

  \end{threeparttable}

  \endgroup
\end{table*}

\section{Conclusions and future research directions}
\label{sec:Conclusions}

In this paper, a systematic, detailed, and thorough comparative evaluation study of recent Deep Learning (DL)-based methods for X-ray object detection was conducted, incorporating six of the most recent, large-scale, and widely used public datasets for X-ray item detection (namely, OPIXray, CLCXray, SIXray, EDS, HiXray, and PIDray) and ten different state-of-art object detection schemes, covering all main categories present in the literature (namely, generic Convolutional Neural Network (CNN), custom (X-ray-specific) CNN, generic transformer and generic hybrid CNN-transformer architectures). Using a comprehensive set of both detection and time/computational-complexity performance metrics, a thorough analysis of the produced results led to the extraction of critical observations and detailed insights, focusing on the following key axes: a) Overall behavior of the various object detection schemes, b) Object-level detection performance investigation, c) Dataset-specific observations, and d) Time efficiency and computational complexity analysis. The fundamental outcome of this study is that there is no single type of detector or class of methods (i.e, CNN, transformer, or hybrid) that is clearly shown advantageous across all benchmarks. To this end, the development of a real-world automated X-ray investigation scheme requires careful consideration of several critical factors, including problem complexity (e.g., degree of object occlusion and presence of clutter), detection robustness/consistency, physical properties of the objects of interest (e.g., material density, geometric complexity, size, etc.), and time performance aspects.

The insights extracted from the current study and the current limitations of the literature suggest at the same time possible future research directions in the field. Among the various pathways, the following considerations are likely to lead to promising outcomes: a) Development of additional, broader, and more challenging/diverse public benchmarks, so as to facilitate the development of robust solutions and rigorous evaluation, b) Design of hybrid CNN-transformer and/or custom (X-ray-specific) architectures that will pay particular attention to the underlying architectural choices (i.e., avoiding possible architectural disharmony occurrences), and c) Development of time-efficient inspection schemes for real-world application scenarios (i.e., emphasizing on architectural and deployment optimization aspects).

\section*{Authorship contribution statement}
Jorgen Cani: Methodology, Software, Validation, Investigation, Data Curation, Writing - Original Draft, Visualization; Christos Diou: Writing - Review \& Editing; Spyridon Evangelatos: Writing - Review \& Editing; Vasileios Argyriou: Writing - Review \& Editing; Panagiotis Radoglou-Grammatikis: Writing - Review \& Editing; Panagiotis Sarigiannidis: Writing - Review \& Editing; Iraklis Varlamis: Writing - Review \& Editing; Georgios Th. Papadopoulos: Conceptualization, Methodology, Resources, Writing - Original Draft, Writing - Review \& Editing, Supervision, Project administration, Funding acquisition.

\section*{Declaration of competing interest}
The authors declare no conflict of interest.

\section*{Acknowledgments}
The research leading to these results has received funding from the European Union’s Horizon Europe research and innovation programme under grant agreement No. 101073876 (Ceasefire). This publication reflects only the authors’ views. The European Union is not liable for any use that may be made of the information contained therein.


\bibliographystyle{elsarticle-harv}
\balance
\bibliography{bibliography}

\end{document}